\documentclass[11pt]{article}

\usepackage{acl}

\usepackage{times}
\usepackage{latexsym}
\usepackage{booktabs}
\usepackage{array}
\usepackage[most]{tcolorbox}
\usepackage{makecell}
\usepackage[table]{xcolor}
\usepackage{colortbl}
\usepackage{capt-of}
\usepackage{url}
\usepackage{caption}
\colorlet{overallgray}{lightgray!20}
\usepackage{afterpage}
\usepackage[T1]{fontenc}

\usepackage[utf8]{inputenc}

\usepackage{microtype}

\usepackage{inconsolata}

\usepackage{graphicx}
\usepackage{subcaption}
\usepackage{float}
\usepackage{multirow}
\usepackage{longtable}
\usepackage{tabularx}
\usepackage{amsmath}
\usepackage{adjustbox}
\usepackage{enumitem}
\tcbuselibrary{raster}

\definecolor{PromptRed}{HTML}{D94A38}
\definecolor{PromptYellow}{HTML}{D89C00}
\definecolor{PromptBlue}{HTML}{2F6FED}
\definecolor{PromptGreen}{HTML}{18A77B}

\newtcolorbox{evalpromptbox}[3]{%
    enhanced,
    colback=#1!5,
    colframe=#1,
    colbacktitle=#1,
    coltitle=white,
    fonttitle=\bfseries,
    fontupper=\ttfamily\scriptsize,
    title={#2},
    height=#3,
    boxrule=0.5pt,
    arc=1mm,
    boxsep=0pt,
    left=1.5mm,
    right=1.5mm,
    top=1mm,
    bottom=1mm,
    before skip=0pt,
    after skip=0pt
}

\title{CapMem: A Benchmark for Caption-Based Episodic Memory in Egocentric Video}

\author{
  \textbf{Dingli Liang}$^{1,*}$,
  \textbf{Yiqiao Xie}$^{2,*}$,
  \textbf{Yukai Huang}$^{3}$,
  \textbf{Zhaokai Wang}$^{4}$,
  \textbf{Weitong Cai}$^{5}$
  \\
  \textbf{Guangwen Feng}$^{7}$,
  \textbf{Jifei Song}$^{6}$,
  \textbf{Zhensong Zhang}$^{6,\dagger}$,
  \textbf{Hang Zhang}$^{7,\dagger}$
  \\[3pt]
  \textsuperscript{1}University College London,
  \textsuperscript{2}Imperial College London,
  \textsuperscript{3}Durham University
  \\
  \textsuperscript{4}Shanghai Jiao Tong University,
  \textsuperscript{5}Queen Mary University of London
  \\
  \textsuperscript{6}Huawei Noah's Ark Lab,
  \textsuperscript{7}Independent Researcher
  \\[3pt]
  {\small $^{*}$Equal contribution.
  $^{\dagger}$Corresponding authors.}
  \\
  {\small
  \href{mailto:miruku.hzhang@gmail.com}{miruku.hzhang@gmail.com},
  \href{mailto:zhangzhensong@huawei.com}{zhangzhensong@huawei.com}}
}

\begin{document}
\maketitle
\begin{abstract}
Wearable assistants require episodic memory over egocentric video, yet current vision-language models face bounded frame budgets, growing visual-token costs, and long-context retrieval failures. Under these practical constraints, we study whether textual captions can serve as reusable episodic memory. We define the Episodic Memory Video Caption QA task and introduce CapMem, a human-annotated benchmark with 75 videos totaling 33.7 hours, and 1,000 multiple-choice questions across 16 scenarios. On long videos ($>20$ min), full-coverage CaptionQA with 30s and 60s caption windows outperforms direct VideoQA for 10/12 and 8/12 models, respectively. On the same video subset, a matched-frame control across six Qwen models retains mean accuracy gains of 3.22 and 2.55 points, respectively. Our caption-guided retrieve-and-verify harness further improves accuracy by up to 5.3 points. These results support the effectiveness of caption memory for episodic reasoning over long egocentric video.
\end{abstract}

\section{Introduction}
Recent advances in vision-language models (VLMs) have turned wearable vision devices from passive recorders into proactive assistants, supporting tasks from vision-based navigation~\cite{tokmurziyev2025llmglasses} to laboratory procedure assistance~\cite{cong2025labos}. Yet such assistants still struggle to retrieve information from continuous egocentric video streams, such as locating a misplaced object or recalling a specific action from hours of visual history~\cite{Baermann_2022_CVPR, mangalam2024egoschema}. This challenge is fundamentally about episodic memory: recovering what happened, where it happened, and when it happened from past experience~\cite{grauman2022ego4d, majumder2024openeqa, chen2024egothink}. Extending this ability to egocentric video stream is challenging: processing continuous visual input is computationally expensive~\cite{song2024moviechat, liu2024worldmodel}, while visual-token budgets limit the number of frames VLMs can process per inference, resulting in sparser temporal coverage and a greater risk of missing relevant evidence ~\cite{lu2025bvllm, tang2025adaptive}. Even with relevant frames, long multimodal contexts can exacerbate fine-grained perception failures~\cite{fu2024blink} and the ``lost-in-the-middle'' effect~\cite{liu2024lost, tian2025identifying, fei2026small} (Figure~\ref{fig:episodic_memory}). An alternative is to convert visual streams into textual memory. Although caption-based representations have shown promise for static images~\cite{yang2025captionqa, xing2025caprl}, whether they can support long-form egocentric episodic memory is largely unexplored.

We address this gap by defining \emph{Episodic Memory Video Caption QA}, a task that evaluates whether self-generated captions can serve as episodic memory in place of direct visual input for egocentric QA. Existing egocentric benchmarks~\cite{grauman2022ego4d, grauman2024egoexo4d, yang2025egolife} are not designed for this purpose: they emphasize retrieval or direct VideoQA and often lack questions probing fine-grained visual details and long-range temporal recall. Moreover, VLM-generated questions~\cite{yang2025captionqa, maaz2023videochatgpt, chen2024sharegpt4video, li2024videovista, zhang2025llavavideo} can inherit the spatial and temporal weaknesses the benchmark seeks to measure. To provide a cleaner testbed, we introduce CapMem, a human-annotated benchmark for egocentric episodic memory. It contains 1,000 human-authored questions grounded in 75 public egocentric videos totaling 33.7 hours across 16 scenarios. CapMem supports three task settings, a two-stage evaluation pipeline with a CleanQA safeguard, and five diagnostic dimensions. Evidence timestamps and fine-grained visual tags enable precise error analysis.
\begin{figure*}[t]
    \centering
    \includegraphics[width=\textwidth,clip,trim=0 50 0 8]{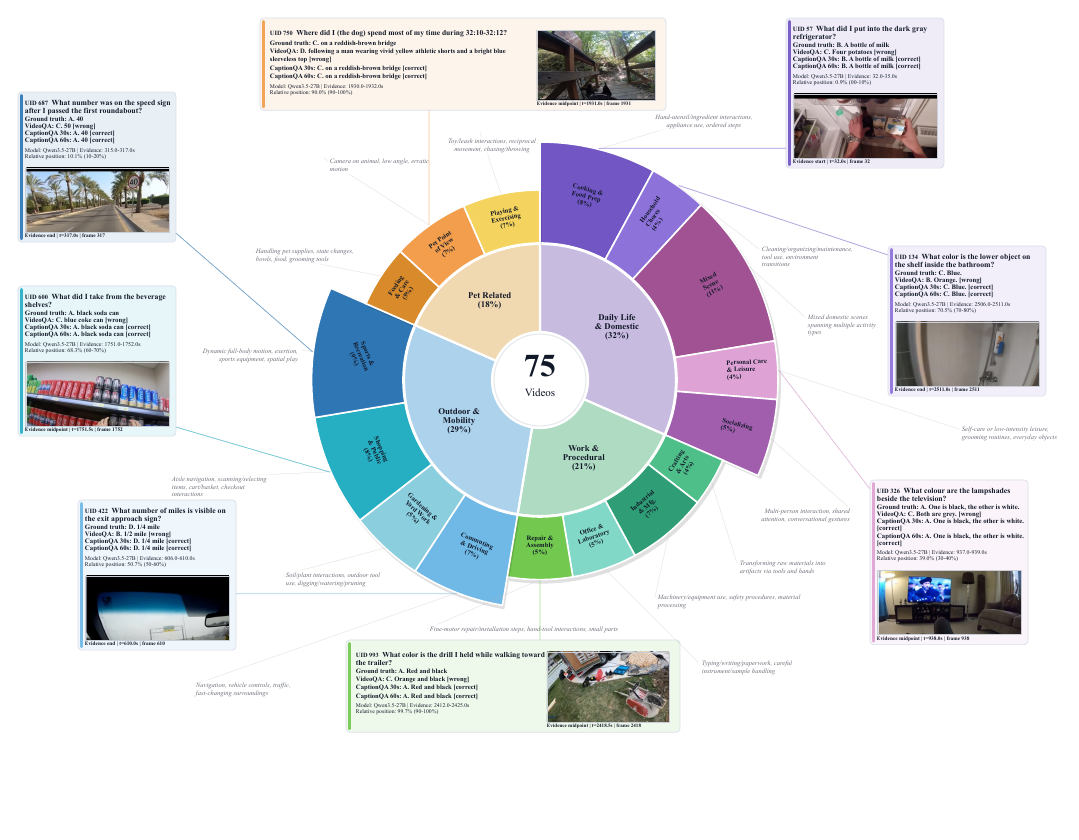}
    \vspace{-13pt}
    \captionsetup{font=footnotesize}
    \caption{CapMem scenario coverage and examples where CaptionQA
succeeds but VideoQA fails, percentage for video share.}
    \label{fig:sunburst_distribution}
    \vspace{-18pt}
\end{figure*}
We also filter out questions that can be answered without visual evidence.

Using CapMem, we evaluate 12 proprietary and open-weight VLMs. Across the full benchmark, full-coverage CaptionQA outperforms direct VideoQA for 8/12 models with 30-second captions and 6/12 with 60-second captions. These gains concentrate on videos longer than 20 minutes, where CaptionQA outperforms direct VideoQA for 10/12 and 8/12 models, while direct VideoQA remains competitive on shorter videos. After matching visual frames, CaptionQA retains mean long-video gains of 3.22 and 2.55 points across six Qwen models, with the clearest statistical evidence in the 30-second setting. Cross-captioning shows that performance depends jointly on caption quality and downstream text reasoning. Caption memory also amortizes repeated visual processing across multiple queries but may omit subtle visual details. We therefore introduce a retrieve-and-verify harness that uses captions as a semantic index and selectively revisits sparse visual evidence, improving most open-weight models. Together, the task, benchmark, controlled analyses, and harness support that caption memory is an effective approach to long-form egocentric episodic reasoning under current VLM constraints.


\section{Related Work}
\subsection{Egocentric Datasets and Benchmarks}

Egocentric vision has progressed from clip-level action recognition~\cite{Damen2020RESCALING} to large-scale datasets~\cite{grauman2022ego4d, grauman2024egoexo4d, engel2023projectaria, yang2025egolife}. Existing assistant-oriented datasets and QA benchmarks, including QAEGO4D~\cite{Baermann_2022_CVPR}, HoloAssist~\cite{kwon2023holoassist}, and EgoSchema~\cite{mangalam2024egoschema}, mainly focus on short or minute-scale video contexts. More recent egocentric QA efforts, such as MM-Ego~\cite{ye2025mmego} and EgoGazeVQA~\cite{peng2025eye}, scale annotation with automatic or MLLM-generated QA pairs, but may inherit generation-induced language bias. In contrast, CapMem provides a human-annotated, long-form video benchmark.



\subsection{Episodic Memory Tasks}
Episodic memory is central to  wearable assistants~\cite{engel2023projectaria, grauman2022ego4d}. Ego4D established this paradigm with memory-oriented tasks such as NLQ, VQ, and MQ~\cite{grauman2022ego4d}. Subsequent work extends this line to egocentric episodic-memory QA~\cite{Baermann_2022_CVPR}, narration-supervised query localization~\cite{ramakrishnan2023naq}, and cross-view memory reasoning~\cite{grauman2024egoexo4d, egoexomem2026}. While these efforts mainly focus on visual retrieval or direct VideoQA, we evaluate whether caption-based memory can support long-form egocentric QA (Appendix \ref{appendixA}).

\subsection{Video Caption and Harness Engineering}
Recent work tackles the memory and token limits of long-horizon videos by abstracting visual content into textual representations. SiLVR \cite{silvr2025} and MR. Video \cite{pang2025mrvideo} use dense clip-level captions for long-context reasoning with LLMs, while LLMVS \cite{llmvs2025} applies frame-level captions to LLM-based video summarization. However, the effectiveness of caption-based reasoning in continuous, long-form egocentric settings remains underexplored. Our task and benchmark fill this gap, providing a dedicated testbed
for, to our knowledge, the first systematic study of how harness
design can enhance caption-centric reasoning.

\begin{figure}[t]
    \centering
    \includegraphics[
        width=\linewidth,
        trim={25pt 125pt 20pt 25pt},
        clip
    ]{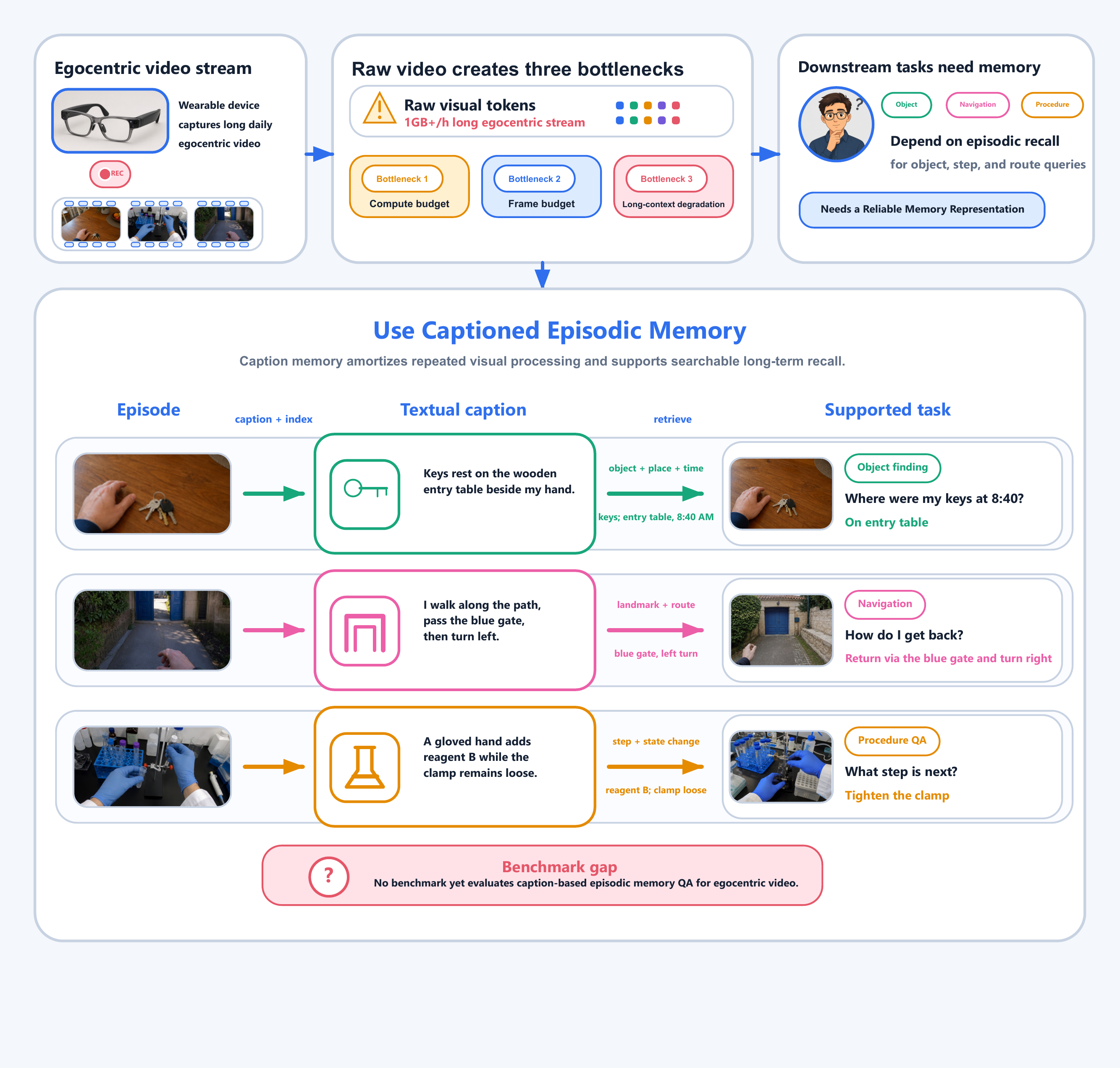}
    \vspace{-12pt}
    \captionsetup{font=footnotesize}
    \caption{Overview of captioned episodic memory for egocentric video QA.}
    \label{fig:episodic_memory}
    \vspace{-15pt}
\end{figure}

\section{Task and Evaluation Setup}

To evaluate the capability of VLMs to use self-generated captions as egocentric episodic memory, we introduce a  QA benchmark, whose evaluation framework features three distinct task settings, a two-stage evaluation pipeline, and a five-dimensional analysis of the capabilities of models.

\subsection{Task Formulation}\label{task}
We define \emph{Episodic Memory Video Caption QA} as a multiple-choice QA task requiring episodic recall, where models answer solely by self-generated timestamped captions from an egocentric video. To assess this capability, we use direct VideoQA as the primary baseline and CleanQA as a safeguard against parametric priors:
\begin{itemize}[label={},leftmargin=0pt,itemindent=0pt,labelsep=0pt,itemsep=1pt,topsep=2pt,parsep=0pt]
    \item \textbf{CleanQA.} The model receives only textual questions and options without visual input, serving as a blind baseline to quantify parametric priors and rule out blind guessing from pre-trained knowledge.
    \item \textbf{Direct VideoQA.} The model answers each question directly from uniformly sampled video frames through its native visual interface, without constructing or using an intermediate textual memory. For brevity, we refer to Direct VideoQA as VideoQA throughout the paper.
    \item \textbf{CaptionQA.} The evaluated model first converts temporally segmented video frames into timestamped captions. It then answers each question using only this textual memory, without access to visual inputs or prior conversation history.
\end{itemize}

\subsection{Benchmark Scope and Statistics}\label{3.2}

A benchmark for evaluating caption-based memory in downstream real-world applications should reflect both the diversity of authentic human routines and the temporal demands of long-horizon video understanding. To this end, CapMem comprises 75 egocentric videos totaling 33.7 hours, spanning four high-value domains: Daily Life \& Domestic, Outdoor \& Mobility, Work \& Procedural, and Pet Related. Collected across diverse cultural backgrounds, the benchmark covers 16 precise professional and domestic scenarios, ranging from industrial machinery repair to outdoor activities, sports, and human--pet interaction (Figure~\ref{fig:sunburst_distribution}). Beyond scenario coverage, the benchmark is intentionally length-challenging: 37.3\% of the videos are short clips ($\leq20$ mins) for verifying foundational video understanding, whereas 62.7\% are long videos ($>20$ mins, up to 1 hour) to stress episodic memory over extended contexts. This emphasis is further reflected in the QA distribution, where 79.2\% of all questions are assigned to the long videos (Table~\ref{fig:figure3_redesign}). In total, the benchmark contains 1,000 multiple-choice questions, averaging 13.3 questions per video. Each question includes four options, one ground-truth answer, and an annotated evidence timestamp indicating where the supporting visual evidence occurs in the video. Table~\ref{tab:cat_qa_ego_exo_4w1h} summarizes the distribution of questions across the four domains and QA taxonomy axes (Sec.~\ref{4.2}), while Table~\ref{fig:figure3_redesign} reports the distribution of video duration, evidence location, and fine-grained visual tags. We categorize evidence by its relative temporal position in the video as beginning (0--20\%), middle (20--80\%), or end (80--100\%). The fine-grained visual tags (Sec.~\ref{4.2}) characterize the visual elements incorporated into the questions and answer options to increase perceptual difficulty.

\vspace{-5pt}
\begingroup
\setlength{\abovecaptionskip}{-1pt}
\setlength{\belowcaptionskip}{6pt}
\setlength{\intextsep}{2pt}
\begin{table}
\centering
\caption{QA statistics. Queries per domain are categorized along two orthogonal axes: subject of inquiry (EGO/EXO) and query intent (4W1H).}
\label{tab:cat_qa_ego_exo_4w1h}
\setlength{\tabcolsep}{6pt}
\renewcommand{\arraystretch}{1.15}
\footnotesize
\resizebox{\linewidth}{!}{%
\begin{tabular}{l r c r @{\hspace{18pt}} r r r r r}
\toprule
\textbf{Category} & \textbf{\#QA} & \textbf{(EGO \,|\, EXO)} & 
\textbf{WHO} & \textbf{WHAT} & \textbf{WHERE} & \textbf{WHEN} & \textbf{HOW} \\
\midrule
Daily Life & 404 & (93\,|\,311)  & 20 & 203 & 54 & 40 & 87 \\
Outdoor & 298 & (49\,|\,249)  & 11 & 151 & 19 & 44 & 73 \\
Pet & 115 & (32\,|\,83)  & 1 & 56 & 16 & 25 & 17 \\
Work & 183 & (50\,|\,133) & 14 & 104 & 8 & 18 & 39 \\
\midrule
\textbf{Total} & \textbf{1000} & \textbf{(224\,|\,776)} & \textbf{46} & \textbf{514} & \textbf{97} & \textbf{127} & \textbf{216} \\
\bottomrule
\end{tabular}%
}
\vspace{-11pt}
\end{table}
\endgroup

\subsection{Evaluation Framework and Dimensions}
We design a two-stage evaluation framework:
\begin{itemize}[label={},leftmargin=0pt,itemindent=0pt,labelsep=0pt,itemsep=2pt,topsep=2pt,parsep=0pt]
    \item \textbf{Stage 1: Baseline Safeguard.} CleanQA is used as a gating test for semantic bias and data contamination. Only models performing near chance level (15\%--35\%) proceed to the main evaluation. To preserve the validity and fairness of downstream comparisons, models falling outside this range are excluded from the subsequent VideoQA--CaptionQA evaluation. Because guessable questions were filtered during benchmark construction (Sec.~\ref{4.3}), unusually high performance at this stage may indicate training-data contamination.

    \item \textbf{Stage 2: Comparative Assessment.} Models that pass the initial gate are evaluated under VideoQA and CaptionQA, as defined in Sec.~\ref{task}. The comparison measures how well self-generated textual memory can substitute for direct visual input in egocentric episodic memory tasks. We analyze the performance along the following dimensions:
\end{itemize}
\begin{itemize}[label={},leftmargin=0pt,itemindent=0pt,labelsep=0pt,itemsep=1pt,topsep=2pt,parsep=0pt]
    \item \textbf{Overall QA Accuracy.} Accuracy over the full question set.
    \item \textbf{Computational Cost.} Average per-question token consumption.
    \item \textbf{Fine-grained Visual Accuracy.} Accuracy reported separately for the four tag groups.
    \item \textbf{Evidence-location QA Accuracy.} Accuracy across the evidence-location groups defined in Sec.~\ref{3.2}, testing sensitivity to the ``lost-in-the-middle'' phenomenon.
    \item \textbf{Duration-wise QA Accuracy.} Accuracy across $\leq20$-minute and $>20$-minute video subsets.
\end{itemize}


\begin{table}
\centering
\small
\setlength{\tabcolsep}{5pt}
\renewcommand{\arraystretch}{1.08}
\caption{Dataset statistics. Fine-grained tags are multi-label.}
\label{fig:figure3_redesign}
\begin{tabular}{llc}
\toprule
\textbf{Aspect} & \textbf{Category} & \textbf{Count} \\
\midrule
\multirow{2}{*}{Video duration}
& $\leq20$ mins videos & 208 Qs (20.8\%) \\
& $>20$ mins videos & 792 Qs (79.2\%) \\
\midrule
\multirow{3}{*}{Evidence location}
& Begin & 348 Qs (34.8\%) \\
& Middle & 507 Qs (50.7\%) \\
& End & 145 Qs (14.5\%) \\
\midrule
\multirow{4}{*}{Fine-grained tag}
& Object & 962 Qs (96.2\%) \\
& Attribute & 708 Qs (70.8\%) \\
& Spatial & 581 Qs (58.1\%) \\
& Action & 844 Qs (84.4\%) \\
\bottomrule
\end{tabular}
\vspace{-20pt}
\end{table}

\section{Benchmark Construction}
\subsection{Video Collection}\label{4.1}
Our video pool is drawn from seven egocentric datasets: Ego4D \cite{grauman2022ego4d}, EgoLife \cite{yang2025egolife}, EgoPet \cite{bar2024egopet}, EPIC-KITCHENS \cite{damen2018scaling}, HoloAssist \cite{kwon2023holoassist}, ENIGMA-51 \cite{ragusa2024enigma51}, and CASTLE2024 \cite{rossetto2025castle}. We release the CapMem annotations and associated metadata for non-commercial research purposes only, without redistributing the source videos or frames; dataset-specific licensing conditions are detailed in Appendix~\ref{app:licensing}.


\subsection{Question Generation}\label{4.2}
We generate QA pairs through a taxonomy-guided annotation pipeline. The taxonomy is defined along two axes: the subject of inquiry, distinguishing questions about the camera wearer and their actions or attributes (Ego) from questions about external entities in view (Exo), and the query intent, organized by the 4W1H forms (Who, What, Where, When, How). Based on this taxonomy, we design 50 question templates covering common episodic reasoning patterns, including object counting, state transitions, and ego--exo interactions (Appendix~\ref{temp}).
For each video, human annotators select suitable templates and instantiate them with concrete entities or events from the video. To increase fine-grained discrimination difficulty, both questions and answer options are enriched with fine-grained cues spanning object identity, visual attributes, spatial relations, and actions, while the options are constructed with distractors that require models to distinguish subtle visual details.

\subsection{Question Filtering}\label{4.3}
After the generation of QA pairs, we adopt a two-stage filtering pipeline to discard flawed data. More quality control detail can be found in Appendix \ref{app:annotation_pipeline}.
\begin{itemize}[label={},leftmargin=0pt,itemindent=0pt,labelsep=0pt,itemsep=1pt,topsep=2pt,parsep=0pt]
    \item \textbf{Step 1: Text-Only Filtering.} VLMs may answer questions without visual reasoning by exploiting linguistic priors, commonsense knowledge, or option-level shortcuts~\cite{li2023evaluating, liu2024mmbench, zhang2026watch}. We therefore run a text-only blind test with a four-model ensemble (GPT-5.2, Gemini 3, Qwen3.5, and InternVL). Questions answered correctly by at least three models are considered visually bypassable and removed. Residual biases are further controlled by the CleanQA gate in our evaluation framework.
    \item \textbf{Step 2: Visual Solvability Filtering.} Human validators verify each query using only the source video. Queries with ambiguous or subjective visual evidence are discarded.
\end{itemize}

\section{Experiment and Discussion}
\subsection{Experiment Setting}  
Our evaluation encompasses 12 models, including representative proprietary models (Gemini 3 Flash~\cite{gemini3flash2025} and GPT-5.2~\cite{openai2025gpt52}) and a comprehensive suite of open-weight models spanning various parameter scales: InternVL3.5~\cite{wang2025internvl35}, Qwen3-VL~\cite{bai2025qwen3vl}, and Qwen3.5~\cite{qwen2026qwen35}. We test the models under distinct settings:
\begin{itemize}[label={},leftmargin=0pt,itemindent=0pt,labelsep=0pt,itemsep=1pt,topsep=2pt,parsep=0pt]
    \item \textbf{CleanQA.} We run strictly single-turn, zero-shot inference. Each query is processed independently with an empty context window, clearing the KV-cache and conversation history to prevent cross-query contamination.
    \item \textbf{VideoQA.} We extract frames at 1 FPS and uniformly subsample when the sequence exceeds context limits. Sequences are capped at 128 frames for InternVL3.5 and 768 frames for Qwen; for cross-model comparability, closed-source APIs are also capped at 768 frames. Models process visual inputs by their native modality interfaces in a single-turn manner. Due to closed-source API budget constraints, Gemini 3 Flash VideoQA is taken on a 786-question subset; all other reported settings use the full benchmark. This setting represents the standard direct video-input interface of general-purpose VLMs under their supported frame budgets. Our goal is to evaluate CaptionQA against this standard baseline, rather than claim superiority over all visual-reasoning pipelines.
    \item \textbf{Full-coverage CaptionQA.} Models deterministically caption the entire 1-FPS frame sequence in consecutive 30- or 60-second windows (temperature = 0), without a video-level frame cap. They concatenate the captions with timestamps as textual episodic memory and answer each query using only this text. Visual inputs are unavailable at QA time, and the captioning and QA prompts are shared across models (Appendix~\ref{prompt})).
\end{itemize}
\noindent\textbf{Frame-aligned Control.} Because CaptionQA may process more frames than VideoQA, we conduct a frame-aligned control on six Qwen models to determine whether its gains arise from caption-based memory rather than greater visual coverage. Both settings use all 1-FPS frames up to 768, or the same 768 uniformly sampled frames otherwise; CaptionQA groups the selected frames by timestamp into 30- or 60-second windows to construct the caption for QA.

\noindent\textbf{Cross-captioning Control.} Because self-caption QA jointly reflects caption-generation quality and text-QA capacity, we conduct a cross-captioning control to decouple these two factors. Captions from six Qwen models are paired with Qwen3.5-2B and Qwen3.5-27B as QA models. We evaluate every captioner--QA pairing using the same frames uniformly sampled over the full videos and capped at 768 per video.

\begin{table}[t]
\centering
\vspace{-8pt}
\small
\setlength{\tabcolsep}{5.5pt}
\renewcommand{\arraystretch}{1.08}
\begin{tabular}{lrrrr}
\toprule
\textbf{Model} & \textbf{Clean} & \textbf{Cap30} & \textbf{Cap60} & \textbf{Video} \\
\midrule
InternVL3.5-8B  & 29.3 & 35.8 & 34.3 & \textbf{39.4} \\
InternVL3.5-38B & 28.4 & 42.1 & 39.8 & \textbf{42.8} \\
Qwen3-VL-2B     & 27.0 & \textbf{36.3} & 35.2 & 35.4 \\
Qwen3-VL-4B     & 26.9 & \textbf{42.3} & 40.3 & 37.1 \\
Qwen3-VL-8B     & 25.8 & \textbf{39.6} & 39.0 & 37.2 \\
Qwen3-VL-32B    & 26.1 & \textbf{47.7} & 44.9 & 42.8 \\
Qwen3.5-2B      & 26.6 & \textbf{37.2} & 34.7 & \textbf{37.2} \\
Qwen3.5-4B      & 27.4 & \textbf{41.8} & 39.6 & 39.6 \\
Qwen3.5-9B      & 29.4 & 42.8 & \textbf{43.7} & 42.3 \\
Qwen3.5-27B     & 30.9 & \textbf{49.3} & 46.9 & 45.0 \\
GPT-5.2         & 25.6 & \textbf{52.0} & 51.4 & 37.3 \\
Gemini 3 Flash & 29.5 & 55.7 & 53.2 & \textbf{58.8}\rlap{$^{*}$} \\
\bottomrule
\end{tabular}
\caption{Overall model accuracy across evaluation formats. $^{*}$ means partial test due to API budget.}
\footnotesize
\label{tab:model_accuracy_by_format}
\vspace{-20pt}
\end{table}
 \subsection{Main Results}
We evaluate full-coverage CaptionQA for overall accuracy, video
duration, amortized visual processing, and evidence position, and use
frame-aligned and cross-captioning controls to decouple the effects of
visual coverage, caption quality, and text-QA capacity.

\paragraph{\textrm{\textbf{Full-coverage CaptionQA is competitive under current VLM frame limits.}}} Table~\ref{tab:model_accuracy_by_format} reports the end-to-end results of full-coverage CaptionQA. CleanQA remains near chance (25.6\%--30.9\%), providing a sanity check. With each model's better caption window, full-coverage CaptionQA outperforms VideoQA for 8 of the 12 models. For example, 30-second CaptionQA improves Qwen3-VL-32B from 42.8\% to 47.7\%, Qwen3.5-27B from 45.0\% to 49.3\%, and GPT-5.2 from 37.3\% to 52.0\%. The 30-second setting also generally outperforms the 60-second setting, suggesting that shorter windows better preserve local details. However, VideoQA remains stronger for both InternVL3.5 models and Gemini 3 Flash. Overall, caption memory provides an effective alternative under current frame and context constraints, but does not universally outperform direct visual input.

\begin{figure}
    \centering
    \vspace{-2pt}
    \includegraphics[
        width=\linewidth,
        trim=0 63 0 38,
        clip
    ]{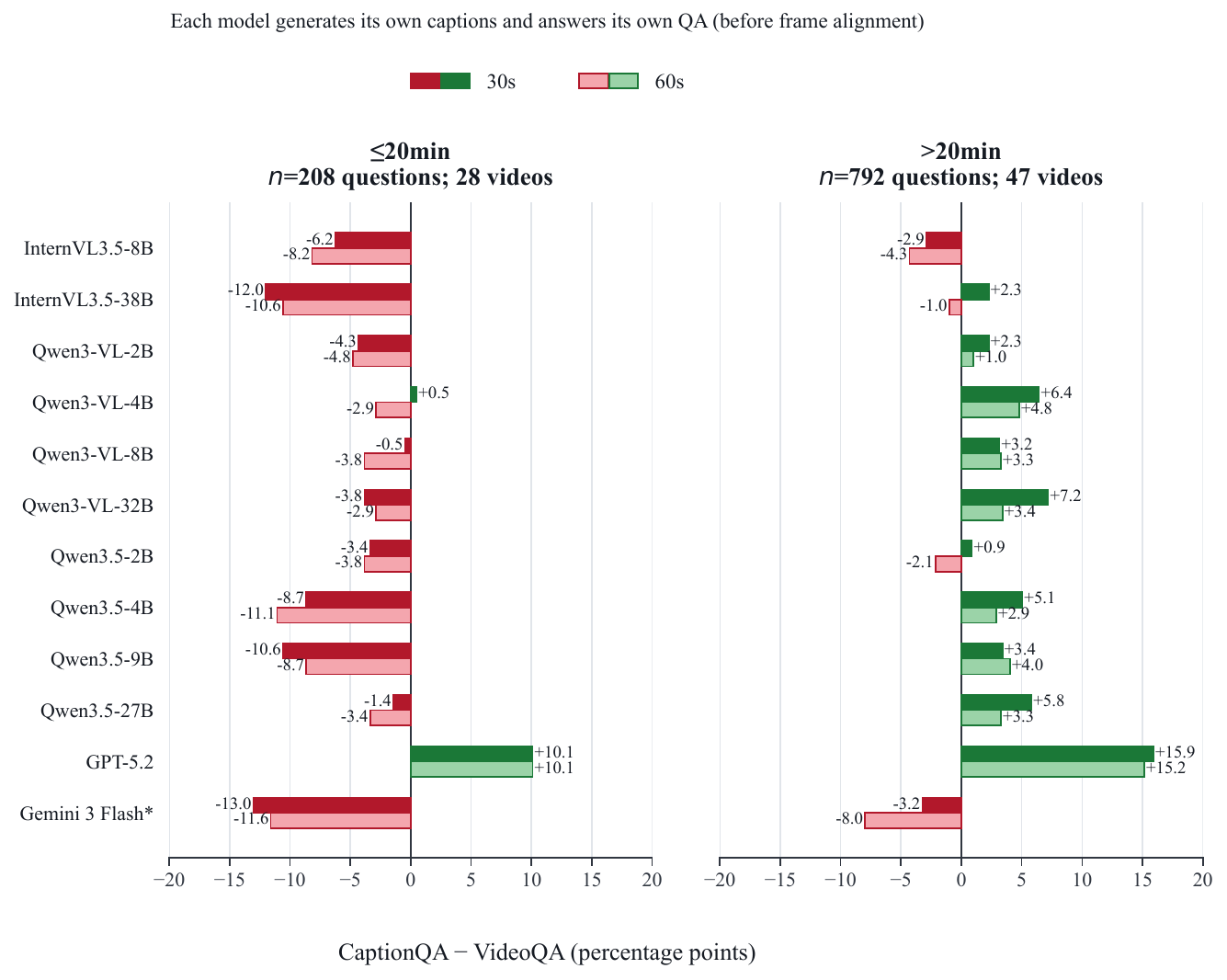}
 
\captionsetup{font=footnotesize}
    \caption{CaptionQA--VideoQA gap by video length before frame alignment.}
    \label{fig:captionqa_advantage_by_video_length}
    \vspace{-20pt}
\end{figure}

\paragraph{\textrm{\textbf{Full-coverage CaptionQA gains concentrate on videos longer than 20 minutes.}}} Figure~\ref{fig:captionqa_advantage_by_video_length} reveals a clear duration-dependent pattern . On videos of at most 20 minutes, CaptionQA generally trails direct VideoQA, indicating that captioning may discard useful visual details when the visual context is not long. On videos longer than 20 minutes, the gaps become positive for most models, reaching +7.2 points for Qwen3-VL-32B with 30-second captions and +15.2 points for GPT-5.2 with 60-second captions. Thus, full-coverage caption memory is particularly useful for long-video reasoning under current VLM frame and context limits, rather than uniformly beneficial across durations.

\begin{figure}
    \centering
    \vspace{-10pt}
    \includegraphics[
        width=\linewidth,
        trim=0 20 0 50,
        clip
    ]{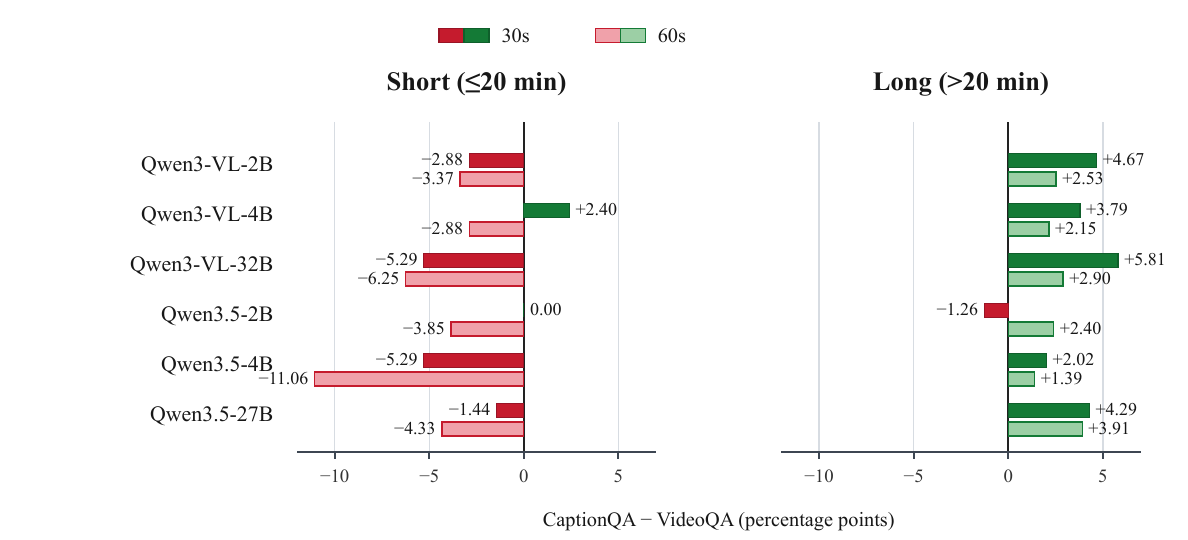}
    \vspace{-13.5pt}
\captionsetup{font=footnotesize}
    \caption{CaptionQA--VideoQA gap by video length after frame alignment.}
    \label{fig1:captionqa_advantage_by_video_length}
    \vspace{-20pt}
\end{figure}

\paragraph{\textrm{\textbf{CaptionQA retains a mean advantage after frame alignment, particularly on long videos.}}}
Table~\ref{tab:aligned_frame_overall} and Figure~\ref{fig1:captionqa_advantage_by_video_length} control for the broader visual coverage of full-coverage CaptionQA. Relative to full coverage, frame alignment lowers mean accuracy by 0.80 points at 30 seconds and leaves it nearly unchanged at 60 seconds (+0.17 points). The resulting mean gaps over direct VideoQA are +2.12 and +0.92 points on the full benchmark, increasing to +3.22 and +2.55 points on the long videos. We estimate uncertainty from question-weighted video-level gaps by 10,000 duration-stratified video-cluster bootstrap samples (Table \ref{tab:aligned_frame_uncertainty}, detail in Appendix \ref{app:uncertainty_estimation}). The 30s confidence interval is positive on long videos, $[+0.45,+6.07]$, whereas the 60s interval is negative  on short videos, $[-9.17,-1.29]$, favoring direct VideoQA. The clearest controlled evidence therefore appears in the 30-second long-video setting. Broader visual coverage contributes to some outcomes but does not fully explain the long-video advantage.

\paragraph{\textrm{\textbf{Caption quality and text-QA capacity both affect CaptionQA.}}}
Table~\ref{tab:cross_captioning} decouples the two capabilities in CaptionQA task. With the captioner fixed, Qwen3.5-27B outperforms Qwen3.5-2B by an average of 8.1 points at 30 seconds and 6.7 points at 60 seconds, showing that a stronger QA model can better use the same captions. With the QA model fixed, varying the captioner makes a max--min accuracy spread of 2.2--6.1 points. Self-captioning is not always optimal: Qwen3.5-2B improves from 36.2\% with its own 30-second captions to 39.8\% with captions from Qwen3-VL-32B. These results show  CaptionQA depends on both memory construction and downstream reasoning.

\begin{table}
    \centering
    \footnotesize
    \setlength{\tabcolsep}{2.2pt}
    \begin{tabular}{lccccc}
        \toprule
        Model & VideoQA & \multicolumn{2}{c}{Cap30} & \multicolumn{2}{c}{Cap60} \\
        \cmidrule(lr){3-4}\cmidrule(lr){5-6}
              &       & Orig. & Align. & Orig. & Align. \\
        \midrule
        Qwen3-VL-2B  & 35.4 & 36.3 & 38.5 & 35.2 & 36.7 \\
        Qwen3-VL-4B  & 37.1 & 42.3 & 40.6 & 40.3 & 38.2 \\
        Qwen3-VL-32B & 42.8 & 47.7 & 46.3 & 44.9 & 43.8 \\
        Qwen3.5-2B   & 37.2 & 37.2 & 36.2 & 34.7 & 38.3 \\
        Qwen3.5-4B   & 39.6 & 41.8 & 40.1 & 39.6 & 38.4 \\
        Qwen3.5-27B  & 45.0 & 49.3 & 48.1 & 46.9 & 47.2 \\
        \midrule
        Mean         & 39.5 & 42.4 & 41.6 & 40.3 & 40.4 \\
        \bottomrule
    \end{tabular}
    \caption{Overall accuracy (\%) before/after frame alignment.}
    \label{tab:aligned_frame_overall}
    
\end{table}

\begin{table}
\centering
\vspace{-20pt}
\footnotesize
\setlength{\tabcolsep}{3.5pt}
\begin{tabular}{lccc}
\toprule
Scope & Int. & Mean & 95\% CI \\
\midrule
Full & 30s & $+2.12$ & $[-0.27,+4.60]$ \\
Full & 60s & $+0.92$ & $[-1.38,+3.18]$ \\
\midrule
$>20$ min & 30s & $+3.22$ & $\mathbf{[+0.45,+6.07]}$ \\
$>20$ min & 60s & $+2.55$ & $[-0.11,+5.23]$ \\
\midrule
$\leq20$ min & 30s & $-2.08$ & $[-6.35,+2.24]$ \\
$\leq20$ min & 60s & $-5.29$ & $\mathbf{[-9.17,-1.29]}$ \\
\bottomrule
\end{tabular}
\caption{Frame-aligned mean gaps and 95\% video-cluster bootstrap confidence intervals.}
\label{tab:aligned_frame_uncertainty}
\end{table}

\begin{table}
\centering
\vspace{-20pt}
\footnotesize
\setlength{\tabcolsep}{3.5pt}
\begin{tabular}{@{}lcccc@{}}
\toprule
& \multicolumn{2}{c}{30s} & \multicolumn{2}{c}{60s} \\
\cmidrule(lr){2-3}\cmidrule(lr){4-5}
Captioner & 2B QA & 27B QA & 2B QA & 27B QA \\
\midrule
Qwen3-VL-2B
& 37.6 & 45.1 & 36.8 & 41.8 \\
Qwen3-VL-4B
& 37.4 & 46.5 & 38.0 & 44.3 \\
Qwen3-VL-32B
& \textbf{39.8} & \textbf{48.9} & 37.4 & 46.6 \\
Qwen3.5-2B
& \underline{36.2} & 42.8 & \underline{38.3} & 42.5 \\
Qwen3.5-4B
& 38.1 & 45.2 & 36.4 & 43.1 \\
Qwen3.5-27B
& 38.8 & \underline{48.1}
& \textbf{38.6} & \textbf{\underline{47.2}} \\
\bottomrule
\end{tabular}
\caption{Cross-captioning accuracy (\%) on the full benchmark; self-captioning results are underlined.}
\label{tab:cross_captioning}
\end{table}

\paragraph{\textrm{\textbf{Caption memory amortizes visual processing across queries.}}}
Table~\ref{tab:token_amortization_all_models} illustrates the reuse
of 30-second caption memory. For $N$ questions sharing one memory,
$C_{\mathrm{Cap}}(N)=(T_{\mathrm{caption}}+
\sum_{i=1}^{N}T_{\mathrm{QA},i})/N$. CaptionQA uses logged
construction and QA tokens, whereas VideoQA uses an analytical
visual-patch proxy (selected frames $\times 1{,}024$ plus text).
Under this accounting, Qwen CaptionQA decreases from 330k--335k
per question at 1Q/video to 14k--20k at 32Q/video. The 
advantage is reusable visual processing: even constructing both
caption granularities requires 101,814 frame inputs, versus 725,973
for Direct VideoQA, reducing repeated visual-frame processing by
$7.1\times$ (Table \ref{tab:frame_accounting}). Realized token and runtime costs remain
processor-dependent (Appendix~\ref{app:compute_accounting}).

\begin{table}
\centering
\vspace{-13pt}
\footnotesize
\setlength{\tabcolsep}{3.2pt}
\renewcommand{\arraystretch}{1.02}
\resizebox{\columnwidth}{!}{%
\begin{tabular}{lrrrrrrr}
\toprule
\textbf{Model} & \multicolumn{6}{c}{\textbf{CaptionQA30s}} & \textbf{VideoQA} \\
\cmidrule(lr){2-7}
 & \textbf{1Q} & \textbf{2Q} & \textbf{4Q} & \textbf{8Q} & \textbf{16Q} & \textbf{32Q} & \\
\midrule
InternVL3.5-8B  & 429 & 217 & 111 & 58  & 32  & 18 & 131 \\
InternVL3.5-38B & 429 & 217 & 111 & 58  & 32  & 18 & 131 \\
Qwen3-VL-2B     & 333 & 170 & 89  & 48  & 28  & 18 & 743 \\
Qwen3-VL-4B     & 333 & 171 & 89  & 48  & 28  & 18 & 743 \\
Qwen3-VL-8B     & 332 & 169 & 88  & 47  & 27  & 16 & 743 \\
Qwen3-VL-32B    & 335 & 172 & 91  & 50  & 30  & 20 & 743 \\
Qwen3.5-2B      & 330 & 167 & 85  & 45  & 24  & 14 & 743 \\
Qwen3.5-4B      & 333 & 170 & 88  & 48  & 27  & 17 & 743 \\
Qwen3.5-9B      & 334 & 171 & 89  & 48  & 28  & 18 & 743 \\
Qwen3.5-27B     & 334 & 171 & 90  & 49  & 29  & 18 & 743 \\
GPT-5.2         & 394 & 200 & 104 & 56  & 31  & 19 & 743 \\
Gemini 3 Flash & 1775 & 891 & 450 & 228 & 118 & 63 & 760\rlap{$^{*}$} \\
\bottomrule
\end{tabular}%
}

\caption{
Amortized per-question workload (k token units) for
30-second CaptionQA: logged input tokens for CaptionQA and an
analytical visual-patch proxy for VideoQA.
}
\label{tab:token_amortization_all_models}
\vspace{-23pt}
\end{table}

\begin{figure}[H]
    \centering
    \vspace{-13pt}
    \includegraphics[width=\linewidth,trim=0 106 0 100,
        clip]{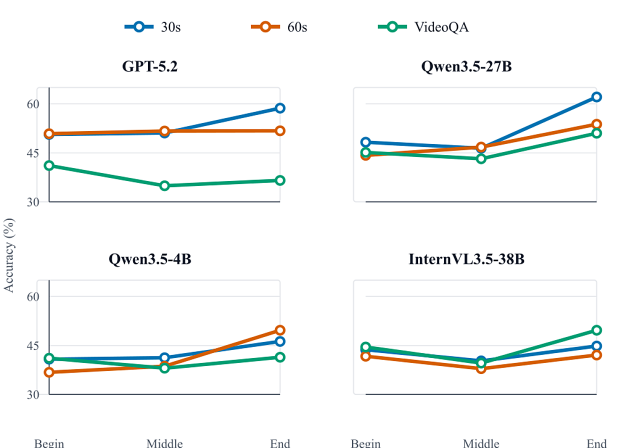}
    \vspace{5pt}
\captionsetup{font=footnotesize}
    \caption{Accuracy by evidence location for four illustrative models spanning different behavior patterns. Complete model-wise results are provided in Figure \ref{fig:accuracy_by_evidence_position1}.}
     \vspace{0pt}
    \label{fig:accuracy_by_evidence_position}
\end{figure}

\paragraph{\textrm{\textbf{Caption memory can reduce middle-position degradation}}} Figure \ref{fig:accuracy_by_evidence_position} shows lower VideoQA accuracy for middle-position evidence in several models, consistent with the ``lost-in-the-middle'' phenomenon. Full-coverage CaptionQA produces a flatter three-bin accuracy profile for several models. For GPT-5.2, it largely removes the observed middle-position drop, while 30-second CaptionQA both reduces the drop and improves accuracy for Qwen3.5-27B. Some InternVL models exhibit flatter profiles despite limited gains in absolute accuracy. Although this pattern is not universal and does not establish a causal mechanism, it suggests that caption memory may make temporally intermediate evidence easier to retrieve for some models. Fine-grained results are provided in Appendix~\ref{app:fine_grained_results}.

\subsection{Harness Engineering}
Although captions are efficient for long-video reasoning, they are a lossy abstraction of the visual stream. Our harness uses captions to localize candidate moments and consults sparse frames only when they can reliably correct a caption-based answer. It has three stages: first, a deterministic weighted retrieval rule selects relevant timestamped caption windows using BM25, hash-based cosine similarity, word overlap, and temporal hints, from which the model produces a retrieved-caption answer. Second, sparse frames from the selected intervals are used to verify which option is visually supported. Third, the baseline CaptionQA answer is kept by default and overridden only when verification returns a valid option with strong support and no explicit clock-time cue, since sparse frames may miss the referenced moment (Appendix~\ref{harness}).
\begin{figure}[t]
    \centering
    \vspace{-18pt}
    \includegraphics[width=\linewidth, trim=0 20 0 10, clip]{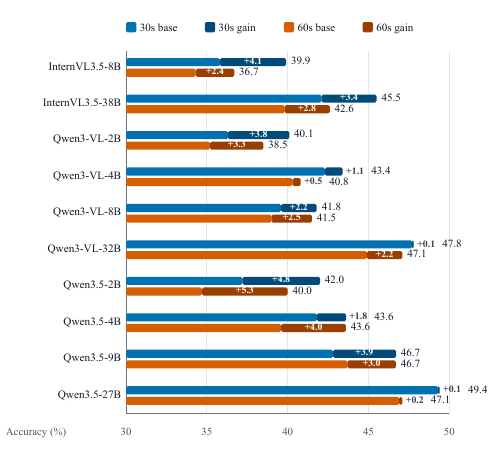}
    \vspace{-15pt}
    \captionsetup{font=footnotesize}
    \caption{Accuracy of open-weight models before and after retrieve-and-verify harnessing.}
    \label{fig:harness_open_models}
    \vspace{-20pt}
\end{figure}
\paragraph{\textrm{\textbf{The harness enhances textual memory baselines.}}}
Figure \ref{fig:harness_open_models} shows consistent gains across all open-weight models and both caption settings. The effect is strongest for smaller models, with Qwen3.5-2B improving by +4.8 at 30s and +5.3 at 60s, while Qwen3.5-27B reaches 49.4\% at 30s. These results show that captions provide useful temporal indices, while sparse frames recover visual details missed by textual memory.

\section{Conclusion}
We introduced CapMem, a human-annotated benchmark for caption-based egocentric episodic memory. Across 12 VLMs, CaptionQA gains concentrate on long videos; under matched frames, the mean gain across 6 selected Qwen models remains positive. Retrieve-and-verify  improves most open-weight models, supporting caption memory as an effective approach under current VLM constraints.

\section*{Limitations}
CapMem contains 75 public egocentric videos and 1,000 QA pairs, providing diverse long-form coverage while necessarily covering only a subset of real-world wearable scenarios. Although videos were selected for scenario coverage rather than
source-dataset balance, Ego4D contributes 56 of 75 videos and 742 of
1,000 questions, so aggregate results may partly reflect its capture
distribution. Our primary visual comparison is limited to standard Direct VideoQA under each model's supported frame budget; we do not evaluate stronger chunked, retrieval-first, or hierarchical visual-memory pipelines. The frame-aligned and cross-captioning controls are restricted to Qwen models, and their controlled effects are heterogeneous, with several confidence intervals including zero, limiting generalization across model families. CaptionQA further depends on caption quality, as missed objects, attributes, or short-lived actions may propagate to downstream QA. Finally, our retrieve-and-verify harness is evaluated as an inference-time strategy rather than an end-to-end trained memory system. Future work should expand the benchmark scale and evaluate broader visual-memory baselines and model families.
\section*{Ethical Considerations}
CapMem is constructed from previously released egocentric video datasets and does not involve collecting new personal recordings. Our annotations are limited to observable events, objects, actions, and spatiotemporal evidence, and avoid inferring identities, sensitive attributes, or personal intentions. Nevertheless, source videos may contain identifiable people and private environments. We therefore do not redistribute source videos or frames; users must obtain the source media separately and comply with the original datasets' licenses, access conditions, and privacy requirements. CapMem releases only its annotations and associated metadata, intended for non-commercial research use, under the license specified in Appendix~\ref{app:licensing}. This license does not supersede any applicable restrictions imposed by the source datasets. Deployment of caption-based memory systems additionally requires appropriate consent, data minimization, and access control.

\bibliography{custom}

@String(CVPR  = {CVPR})

@String(ICCV  = {ICCV})

@String(ECCV  = {ECCV})

@String(CVPRW = {CVPRW})

@article{bai2025qwen3vl,
  title = {{Qwen3-VL} Technical Report},
  author = {Shuai Bai and Yuxuan Cai and Ruizhe Chen and Keqin Chen and Xionghui Chen and Zesen Cheng and Lianghao Deng and Wei Ding and Chang Gao and Chunjiang Ge and Wenbin Ge and Zhifang Guo and Qidong Huang and Jie Huang and Fei Huang and Binyuan Hui and Shutong Jiang and Zhaohai Li and Mingsheng Li and Mei Li and Kaixin Li and Zicheng Lin and Junyang Lin and Xuejing Liu and Jiawei Liu and Chenglong Liu and Yang Liu and Dayiheng Liu and Shixuan Liu and Dunjie Lu and Ruilin Luo and Chenxu Lv and Rui Men and Lingchen Meng and Xuancheng Ren and Xingzhang Ren and Sibo Song and Yuchong Sun and Jun Tang and Jianhong Tu and Jianqiang Wan and Peng Wang and Pengfei Wang and Qiuyue Wang and Yuxuan Wang and Tianbao Xie and Yiheng Xu and Haiyang Xu and Jin Xu and Zhibo Yang and Mingkun Yang and Jianxin Yang and An Yang and Bowen Yu and Fei Zhang and Hang Zhang and Xi Zhang and Bo Zheng and Humen Zhong and Jingren Zhou and Fan Zhou and Jing Zhou and Yuanzhi Zhu and Ke Zhu},
  journal = {arXiv preprint arXiv:2511.21631},
  year = {2025},
  eprint = {2511.21631},
  archiveprefix = {arXiv},
  primaryclass = {cs.CV},
  url = {https://arxiv.org/abs/2511.21631},
}

@inproceedings{bar2024egopet,
  title = {{EgoPet}: Egomotion and Interaction Data from an Animal's Perspective},
  author = {Bar, Amir and Bakhtiar, Arya and Tran, Danny and Loquercio, Antonio and Rajasegaran, Jathushan and LeCun, Yann and Globerson, Amir and Darrell, Trevor},
  booktitle = {Computer Vision -- {ECCV} 2024},
  pages = {377--394},
  publisher = {Springer},
  year = {2024},
  doi = {10.1007/978-3-031-72913-3_21},
  url = {https://doi.org/10.1007/978-3-031-72913-3_21},
}

@inproceedings{Baermann_2022_CVPR,
  title = {Where Did I Leave My Keys? -- Episodic-Memory-Based Question Answering on Egocentric Videos},
  author = {B\"armann, Leonard and Waibel, Alex},
  booktitle = {Proceedings of the IEEE/CVF Conference on Computer Vision and Pattern Recognition Workshops ({CVPRW})},
  pages = {1560--1568},
  month = {June},
  year = {2022},
  doi = {10.1109/CVPRW56347.2022.00162},
  url = {https://doi.org/10.1109/CVPRW56347.2022.00162},
}

@book{bell2009total,
  title = {Total Recall: How the E-Memory Revolution Will Change Everything},
  author = {Bell, Gordon and Gemmell, Jim},
  publisher = {Dutton},
  address = {New York},
  year = {2009},
}

@inproceedings{chen2024sharegpt4video,
  title = {{ShareGPT4Video}: Improving Video Understanding and Generation with Better Captions},
  author = {Lin Chen and Xilin Wei and Jinsong Li and Xiaoyi Dong and Pan Zhang and Yuhang Zang and Zehui Chen and Haodong Duan and Bin Lin and Zhenyu Tang and Li Yuan and Yu Qiao and Dahua Lin and Feng Zhao and Jiaqi Wang},
  booktitle = {Advances in Neural Information Processing Systems},
  volume = {37},
  pages = {19472--19495},
  year = {2024},
  doi = {10.52202/079017-0614},
  url = {https://doi.org/10.52202/079017-0614},
}

@inproceedings{chen2024egothink,
  title = {{EgoThink}: Evaluating First-Person Perspective Thinking Capability of Vision-Language Models},
  author = {Sijie Cheng and Zhicheng Guo and Jingwen Wu and Kechen Fang and Peng Li and Huaping Liu and Yang Liu},
  booktitle = {Proceedings of the IEEE/CVF Conference on Computer Vision and Pattern Recognition ({CVPR})},
  pages = {14291--14302},
  year = {2024},
  doi = {10.1109/CVPR52733.2024.01355},
  url = {https://openaccess.thecvf.com/content/CVPR2024/html/Cheng_EgoThink_Evaluating_First-Person_Perspective_Thinking_Capability_of_Vision-Language_Models_CVPR_2024_paper.html},
}

@article{cong2025labos,
  title = {{LabOS}: The {AI-XR} Co-Scientist That Sees and Works With Humans},
  author = {Le Cong and David Smerkous and Xiaotong Wang and Di Yin and Zaixi Zhang and Ruofan Jin and Yinkai Wang and Michal Gerasimiuk and Ravi K. Dinesh and Alex Smerkous and Lihan Shi and Joy Zheng and Ian Lam and Xuekun Wu and Shilong Liu and Peishan Li and Yi Zhu and Ning Zhao and Meenal Parakh and Simran Serrao and Imran A. Mohammad and Chao-Yeh Chen and Xiufeng Xie and Tiffany Chen and David Weinstein and Greg Barbone and Belgin Caglar and John B. Sunwoo and Fuxin Li and Jia Deng and Joseph C. Wu and Sanfeng Wu and Mengdi Wang},
  journal = {arXiv preprint arXiv:2510.14861},
  year = {2025},
  eprint = {2510.14861},
  archiveprefix = {arXiv},
  primaryclass = {cs.AI},
  url = {https://arxiv.org/abs/2510.14861},
}

@article{conway2000construction,
  title = {The Construction of Autobiographical Memories in the Self-Memory System},
  author = {Conway, Martin A. and Pleydell-Pearce, Christopher W.},
  journal = {Psychological Review},
  volume = {107},
  number = {2},
  pages = {261--288},
  year = {2000},
  doi = {10.1037/0033-295X.107.2.261},
  url = {https://doi.org/10.1037/0033-295X.107.2.261},
}

@article{Damen2020RESCALING,
  title = {Rescaling Egocentric Vision: Collection, Pipeline and Challenges for {EPIC-KITCHENS-100}},
  author = {Dima Damen and Hazel Doughty and Giovanni Maria Farinella and Antonino Furnari and Evangelos Kazakos and Jian Ma and Davide Moltisanti and Jonathan Munro and Toby Perrett and Will Price and Michael Wray},
  journal = {International Journal of Computer Vision},
  volume = {130},
  number = {1},
  pages = {33--55},
  year = {2022},
  doi = {10.1007/s11263-021-01531-2},
  url = {https://doi.org/10.1007/s11263-021-01531-2},
}

@inproceedings{damen2018scaling,
  title = {Scaling Egocentric Vision: The {EPIC-KITCHENS} Dataset},
  author = {Damen, Dima and Doughty, Hazel and Farinella, Giovanni Maria and Fidler, Sanja and Furnari, Antonino and Kazakos, Evangelos and Moltisanti, Davide and Munro, Jonathan and Perrett, Toby and Price, Will and Wray, Michael},
  booktitle = {Proceedings of the European Conference on Computer Vision ({ECCV})},
  pages = {720--736},
  year = {2018},
}

@inproceedings{datta2022episodic,
  title = {Episodic Memory Question Answering},
  author = {Datta, Samyak and Dharur, Sameer and Cartillier, Vincent and Desai, Ruta and Khanna, Mukul and Batra, Dhruv and Parikh, Devi},
  booktitle = {Proceedings of the IEEE/CVF Conference on Computer Vision and Pattern Recognition ({CVPR})},
  pages = {19097--19106},
  year = {2022},
  doi = {10.1109/CVPR52688.2022.01853},
  url = {https://openaccess.thecvf.com/content/CVPR2022/html/Datta_Episodic_Memory_Question_Answering_CVPR_2022_paper.html},
}

@article{doherty2012sensecam,
  title = {Experiences of Aiding Autobiographical Memory Using the {SenseCam}},
  author = {Doherty, Aiden R. and Pauly-Takacs, Katalin and Caprani, Niamh and Gurrin, Cathal and Moulin, Chris J. A. and O'Connor, Noel E. and Smeaton, Alan F.},
  journal = {Human--Computer Interaction},
  volume = {27},
  number = {1--2},
  pages = {151--174},
  year = {2012},
  doi = {10.1080/07370024.2012.656050},
  url = {https://doi.org/10.1080/07370024.2012.656050},
}

@article{engel2023projectaria,
  title = {Project Aria: A New Tool for Egocentric Multi-Modal {AI} Research},
  author = {Jakob Engel and Kiran Somasundaram and Michael Goesele and Albert Sun and Alexander Gamino and Andrew Turner and Arjang Talattof and Arnie Yuan and Bilal Souti and Brighid Meredith and Cheng Peng and Chris Sweeney and Cole Wilson and Dan Barnes and Daniel DeTone and David Caruso and Derek Valleroy and Dinesh Ginjupalli and Duncan Frost and Edward Miller and Elias Mueggler and Evgeniy Oleinik and Fan Zhang and Guruprasad Somasundaram and Gustavo Solaira and Harry Lanaras and Henry Howard-Jenkins and Huixuan Tang and Hyo Jin Kim and Jaime Rivera and Ji Luo and Jing Dong and Julian Straub and Kevin Bailey and Kevin Eckenhoff and Lingni Ma and Luis Pesqueira and Mark Schwesinger and Maurizio Monge and Nan Yang and Nick Charron and Nikhil Raina and Omkar Parkhi and Peter Borschowa and Pierre Moulon and Prince Gupta and Raul Mur-Artal and Robbie Pennington and Sachin Kulkarni and Sagar Miglani and Santosh Gondi and Saransh Solanki and Sean Diener and Shangyi Cheng and Simon Green and Steve Saarinen and Suvam Patra and Tassos Mourikis and Thomas Whelan and Tripti Singh and Vasileios Balntas and Vijay Baiyya and Wilson Dreewes and Xiaqing Pan and Yang Lou and Yipu Zhao and Yusuf Mansour and Yuyang Zou and Zhaoyang Lv and Zijian Wang and Mingfei Yan and Carl Ren and Renzo De Nardi and Richard Newcombe},
  journal = {arXiv preprint arXiv:2308.13561},
  year = {2023},
  eprint = {2308.13561},
  archiveprefix = {arXiv},
  primaryclass = {cs.HC},
  url = {https://arxiv.org/abs/2308.13561},
}

@article{fei2026small,
  title = {Small Vision-Language Models Are Smart Compressors for Long Video Understanding},
  author = {Junjie Fei and Jun Chen and Zechun Liu and Yunyang Xiong and Chong Zhou and Wei Wen and Junlin Han and Mingchen Zhuge and Saksham Suri and Qi Qian and Shuming Liu and Lemeng Wu and Raghuraman Krishnamoorthi and Vikas Chandra and Mohamed Elhoseiny and Chenchen Zhu},
  journal = {arXiv preprint arXiv:2604.08120},
  year = {2026},
  eprint = {2604.08120},
  archiveprefix = {arXiv},
  primaryclass = {cs.CV},
  url = {https://arxiv.org/abs/2604.08120},
}

@inproceedings{fu2024blink,
  title = {{BLINK}: Multimodal Large Language Models Can See but Not Perceive},
  author = {Xingyu Fu and Yushi Hu and Bangzheng Li and Yu Feng and Haoyu Wang and Xudong Lin and Dan Roth and Noah A. Smith and Wei-Chiu Ma and Ranjay Krishna},
  booktitle = {Computer Vision -- {ECCV} 2024},
  pages = {148--166},
  publisher = {Springer},
  year = {2024},
  doi = {10.1007/978-3-031-73337-6_9},
  url = {https://www.ecva.net/papers/eccv_2024/papers_ECCV/html/3356_ECCV_2024_paper.php},
}

@article{gemmell2006mylifebits,
  title = {{MyLifeBits}: A Personal Database for Everything},
  author = {Gemmell, Jim and Bell, Gordon and Lueder, Roger and Drucker, Steven and Wong, Curtis},
  journal = {Communications of the ACM},
  volume = {49},
  number = {1},
  pages = {88--95},
  year = {2006},
  doi = {10.1145/1107458.1107460},
  url = {https://doi.org/10.1145/1107458.1107460},
}

@misc{gemini3flash2025,
  title = {Gemini 3 Flash},
  author = {{Google DeepMind}},
  year = {2025},
  howpublished = {\url{https://ai.google.dev/gemini-api/docs/models}},
  note = {Accessed: 2026-06-17},
}

@inproceedings{grauman2022ego4d,
  title = {{Ego4D}: Around the World in 3,000 Hours of Egocentric Video},
  author = {Kristen Grauman and Andrew Westbury and Eugene Byrne and Zachary Chavis and Antonino Furnari and Rohit Girdhar and Jackson Hamburger and Hao Jiang and Miao Liu and Xingyu Liu and Miguel Martin and Tushar Nagarajan and Ilija Radosavovic and Santhosh Kumar Ramakrishnan and Fiona Ryan and Jayant Sharma and Michael Wray and Mengmeng Xu and Eric Zhongcong Xu and Chen Zhao and Siddhant Bansal and Dhruv Batra and Vincent Cartillier and Sean Crane and Tien Do and Morrie Doulaty and Akshay Erapalli and Christoph Feichtenhofer and Adriano Fragomeni and Qichen Fu and Abrham Gebreselasie and Cristina Gonzalez and James Hillis and Xuhua Huang and Yifei Huang and Wenqi Jia and Weslie Khoo and Jachym Kolar and Satwik Kottur and Anurag Kumar and Federico Landini and Chao Li and Yanghao Li and Zhenqiang Li and Karttikeya Mangalam and Raghava Modhugu and Jonathan Munro and Tullie Murrell and Takumi Nishiyasu and Will Price and Paola Ruiz Puentes and Merey Ramazanova and Leda Sari and Kiran Somasundaram and Audrey Southerland and Yusuke Sugano and Ruijie Tao and Minh Vo and Yuchen Wang and Xindi Wu and Takuma Yagi and Ziwei Zhao and Yunyi Zhu and Pablo Arbelaez and David Crandall and Dima Damen and Giovanni Maria Farinella and Christian Fuegen and Bernard Ghanem and Vamsi Krishna Ithapu and C. V. Jawahar and Hanbyul Joo and Kris Kitani and Haizhou Li and Richard Newcombe and Aude Oliva and Hyun Soo Park and James M. Rehg and Yoichi Sato and Jianbo Shi and Mike Zheng Shou and Antonio Torralba and Lorenzo Torresani and Mingfei Yan and Jitendra Malik},
  booktitle = {Proceedings of the IEEE/CVF Conference on Computer Vision and Pattern Recognition ({CVPR})},
  pages = {18995--19012},
  year = {2022},
  doi = {10.1109/CVPR52688.2022.01842},
  url = {https://openaccess.thecvf.com/content/CVPR2022/html/Grauman_Ego4D_Around_the_World_in_3000_Hours_of_Egocentric_Video_CVPR_2022_paper.html},
}

@article{grauman2024egoexo4d,
  title = {{Ego-Exo4D}: Understanding Skilled Human Activity from First- and Third-Person Perspectives},
  author = {Kristen Grauman and Andrew Westbury and Lorenzo Torresani and Kris Kitani and Jitendra Malik and Triantafyllos Afouras and Kumar Ashutosh and Vijay Baiyya and Siddhant Bansal and Bikram Boote and Eugene Byrne and Zach Chavis and Joya Chen and Feng Cheng and Fu-Jen Chu and Sean Crane and Avijit Dasgupta and Jing Dong and Maria Escobar and Cristhian Forigua and Abrham Gebreselasie and Sanjay Haresh and Jing Huang and Md Mohaiminul Islam and Suyog Jain and Rawal Khirodkar and Devansh Kukreja and Kevin J. Liang and Jia-Wei Liu and Sagnik Majumder and Yongsen Mao and Miguel Martin and Effrosyni Mavroudi and Tushar Nagarajan and Francesco Ragusa and Santhosh Kumar Ramakrishnan and Luigi Seminara and Arjun Somayazulu and Yale Song and Shan Su and Zihui Xue and Edward Zhang and Jinxu Zhang and Angela Castillo and Changan Chen and Xinzhu Fu and Ryosuke Furuta and Cristina González and Prince Gupta and Jiabo Hu and Yifei Huang and Yiming Huang and Weslie Khoo and Anush Kumar and Robert Kuo and Sach Lakhavani and Miao Liu and Mi Luo and Zhengyi Luo and Brighid Meredith and Austin Miller and Oluwatumininu Oguntola and Xiaqing Pan and Penny Peng and Shraman Pramanick and Merey Ramazanova and Fiona Ryan and Wei Shan and Kiran Somasundaram and Chenan Song and Audrey Southerland and Masatoshi Tateno and Huiyu Wang and Yuchen Wang and Takuma Yagi and Mingfei Yan and Xitong Yang and Zecheng Yu and Shengxin Cindy Zha and Chen Zhao and Ziwei Zhao and Zhifan Zhu and Jeff Zhuo and Pablo Arbeláez and Gedas Bertasius and David Crandall and Dima Damen and Jakob Engel and Giovanni Maria Farinella and Antonino Furnari and Bernard Ghanem and Judy Hoffman and C. V. Jawahar and Richard Newcombe and Hyun Soo Park and James M. Rehg and Yoichi Sato and Manolis Savva and Jianbo Shi and Mike Zheng Shou and Michael Wray},
  journal = {International Journal of Computer Vision},
  volume = {133},
  number = {12},
  pages = {8356--8435},
  year = {2025},
  doi = {10.1007/s11263-025-02557-6},
  url = {https://doi.org/10.1007/s11263-025-02557-6},
}

@inproceedings{hodges2006sensecam,
  title = {{SenseCam}: A Retrospective Memory Aid},
  author = {Hodges, Steve and Williams, Lyndsay and Berry, Emma and Izadi, Shahram and Srinivasan, James and Butler, Alex and Smyth, Gavin and Kapur, Narinder and Wood, Ken},
  booktitle = {UbiComp 2006: Ubiquitous Computing},
  series = {Lecture Notes in Computer Science},
  volume = {4206},
  pages = {177--193},
  publisher = {Springer},
  year = {2006},
  doi = {10.1007/11853565_11},
  url = {https://doi.org/10.1007/11853565_11},
}

@inproceedings{kwon2023holoassist,
  title = {{HoloAssist}: An Egocentric Human Interaction Dataset for Interactive {AI} Assistants in the Real World},
  author = {Xin Wang and Taein Kwon and Mahdi Rad and Bowen Pan and Ishani Chakraborty and Sean Andrist and Dan Bohus and Ashley Feniello and Bugra Tekin and Felipe Vieira Frujeri and Neel Joshi and Marc Pollefeys},
  booktitle = {Proceedings of the IEEE/CVF International Conference on Computer Vision ({ICCV})},
  pages = {20270--20281},
  year = {2023},
  doi = {10.1109/ICCV51070.2023.01854},
  url = {https://doi.org/10.1109/ICCV51070.2023.01854},
}

@inproceedings{llmvs2025,
  title = {Video Summarization with Large Language Models},
  author = {Lee, Min Jung and Gong, Dayoung and Cho, Minsu},
  booktitle = {Proceedings of the IEEE/CVF Conference on Computer Vision and Pattern Recognition ({CVPR})},
  pages = {18981--18991},
  year = {2025},
  doi = {10.1109/CVPR52734.2025.01768},
  url = {https://openaccess.thecvf.com/content/CVPR2025/html/Lee_Video_Summarization_with_Large_Language_Models_CVPR_2025_paper.html},
}

@article{li2024videovista,
  title = {{VideoVista}: A Versatile Benchmark for Video Understanding and Reasoning},
  author = {Yunxin Li and Xinyu Chen and Baotian Hu and Longyue Wang and Haoyuan Shi and Min Zhang},
  journal = {arXiv preprint arXiv:2406.11303},
  year = {2024},
  eprint = {2406.11303},
  archiveprefix = {arXiv},
  primaryclass = {cs.CV},
  url = {https://arxiv.org/abs/2406.11303},
}

@inproceedings{li2023evaluating,
  title = {Evaluating Object Hallucination in Large Vision-Language Models},
  author = {Li, Yifan and Du, Yifan and Zhou, Kun and Wang, Jinpeng and Zhao, Wayne Xin and Wen, Ji-Rong},
  booktitle = {Proceedings of the 2023 Conference on Empirical Methods in Natural Language Processing},
  pages = {292--305},
  year = {2023},
  doi = {10.18653/v1/2023.emnlp-main.20},
  url = {https://aclanthology.org/2023.emnlp-main.20/},
}

@article{liu2024worldmodel,
  title = {World Model on Million-Length Video and Language With Blockwise {RingAttention}},
  author = {Hao Liu and Wilson Yan and Matei Zaharia and Pieter Abbeel},
  journal = {arXiv preprint arXiv:2402.08268},
  year = {2024},
  eprint = {2402.08268},
  archiveprefix = {arXiv},
  primaryclass = {cs.LG},
  url = {https://arxiv.org/abs/2402.08268},
}

@article{liu2024lost,
  title = {Lost in the Middle: How Language Models Use Long Contexts},
  author = {Nelson F. Liu and Kevin Lin and John Hewitt and Ashwin Paranjape and Michele Bevilacqua and Fabio Petroni and Percy Liang},
  journal = {Transactions of the Association for Computational Linguistics},
  volume = {12},
  pages = {157--173},
  year = {2024},
  doi = {10.1162/tacl_a_00638},
  url = {https://aclanthology.org/2024.tacl-1.9/},
}

@article{egoexomem2026,
  title = {{EgoExoMem}: Cross-View Memory Reasoning over Synchronized Egocentric and Exocentric Videos},
  author = {Ruiping Liu and Junwei Zheng and Yufan Chen and Di Wen and Shaofang Quan and Chengzhi Wu and Jiaming Zhang and Kailun Yang and Kunyu Peng and Rainer Stiefelhagen},
  journal = {arXiv preprint arXiv:2605.18734},
  year = {2026},
  eprint = {2605.18734},
  archiveprefix = {arXiv},
  primaryclass = {cs.CV},
  url = {https://arxiv.org/abs/2605.18734},
}

@inproceedings{liu2024mmbench,
  title = {{MMBench}: Is Your Multi-modal Model an All-around Player?},
  author = {Liu, Yuan and Duan, Haodong and Zhang, Yuanhan and Li, Bo and Zhang, Songyang and Zhao, Wei and Yuan, Yike and Wang, Jiaqi and He, Conghui and Liu, Ziwei and Chen, Kai and Lin, Dahua},
  booktitle = {Computer Vision -- {ECCV} 2024},
  pages = {216--233},
  publisher = {Springer},
  year = {2024},
  doi = {10.1007/978-3-031-72658-3_13},
  url = {https://doi.org/10.1007/978-3-031-72658-3_13},
}

@inproceedings{lu2025bvllm,
  title = {{B-VLLM}: A Vision Large Language Model with Balanced Spatio-Temporal Tokens},
  author = {Lu, Zhuqiang and Yin, Zhenfei and He, Mengwei and Wang, Zhihui and Liu, Zicheng and Wang, Zhiyong and Hu, Kun},
  booktitle = {Proceedings of the IEEE/CVF International Conference on Computer Vision ({ICCV})},
  pages = {24549--24558},
  year = {2025},
  doi = {10.1109/ICCV51701.2025.02276},
  url = {https://openaccess.thecvf.com/content/ICCV2025/html/Lu_B-VLLM_A_Vision_Large_Language_Model_with_Balanced_Spatio-Temporal_Tokens_ICCV_2025_paper.html},
}

@inproceedings{maaz2023videochatgpt,
  title = {Video-ChatGPT: Towards Detailed Video Understanding via Large Vision and Language Models},
  author = {Maaz, Muhammad and Rasheed, Hanoona and Khan, Salman and Khan, Fahad Shahbaz},
  booktitle = {Proceedings of the 62nd Annual Meeting of the Association for Computational Linguistics (Volume 1: Long Papers)},
  pages = {12585--12602},
  year = {2024},
  doi = {10.18653/v1/2024.acl-long.679},
  url = {https://aclanthology.org/2024.acl-long.679/},
}

@inproceedings{majumder2024openeqa,
  title = {{OpenEQA}: Embodied Question Answering in the Era of Foundation Models},
  author = {Arjun Majumdar and Anurag Ajay and Xiaohan Zhang and Pranav Putta and Sriram Yenamandra and Mikael Henaff and Sneha Silwal and Paul Mcvay and Oleksandr Maksymets and Sergio Arnaud and Karmesh Yadav and Qiyang Li and Ben Newman and Mohit Sharma and Vincent Berges and Shiqi Zhang and Pulkit Agrawal and Yonatan Bisk and Dhruv Batra and Mrinal Kalakrishnan and Franziska Meier and Chris Paxton and Alexander Sax and Aravind Rajeswaran},
  booktitle = {Proceedings of the IEEE/CVF Conference on Computer Vision and Pattern Recognition ({CVPR})},
  pages = {16488--16498},
  year = {2024},
  doi = {10.1109/CVPR52733.2024.01560},
  url = {https://openaccess.thecvf.com/content/CVPR2024/html/Majumdar_OpenEQA_Embodied_Question_Answering_in_the_Era_of_Foundation_Models_CVPR_2024_paper.html},
}

@inproceedings{mangalam2024egoschema,
  title = {{EgoSchema}: A Diagnostic Benchmark for Very Long-Form Video Language Understanding},
  author = {Karttikeya Mangalam and Raiymbek Akshulakov and Jitendra Malik},
  booktitle = {Advances in Neural Information Processing Systems},
  volume = {36},
  pages = {46212--46244},
  year = {2023},
  doi = {10.52202/075280-2004},
  url = {https://proceedings.neurips.cc/paper_files/paper/2023/hash/90ce332aff156b910b002ce4e6880dec-Abstract-Datasets_and_Benchmarks.html},
}

@misc{openai2025gpt52,
  title = {{GPT-5.2}},
  author = {{OpenAI}},
  year = {2025},
  howpublished = {\url{https://openai.com/index/introducing-gpt-5-2/}},
  note = {Accessed: 2026-06-17},
}

@inproceedings{pang2025mrvideo,
  title = {{MR. Video}: {MapReduce} as an Effective Principle for Long Video Understanding},
  author = {Pang, Ziqi and Wang, Yu-Xiong},
  booktitle = {Advances in Neural Information Processing Systems},
  volume = {38},
  pages = {7428--7460},
  year = {2025},
  doi = {10.52202/085713-0231},
  url = {https://proceedings.neurips.cc/paper_files/paper/2025/hash/0a02c2bc2e2148b803c4ade1d71e1d25-Abstract-Conference.html},
}

@inproceedings{peng2025eye,
  title = {In the Eye of MLLM: Benchmarking Egocentric Video Intent Understanding with Gaze-Guided Prompting},
  author = {Peng, Taiying and Hua, Jiacheng and Liu, Miao and Lu, Feng},
  booktitle = {Advances in Neural Information Processing Systems},
  volume = {38},
  pages = {121404--121424},
  year = {2025},
  doi = {10.52202/085713-3659},
  url = {https://doi.org/10.52202/085713-3659},
  note = {Datasets and Benchmarks Track},
}

@misc{qwen2026qwen35,
  title = {{Qwen3.5}},
  author = {{Qwen Team}},
  year = {2026},
  howpublished = {\url{https://huggingface.co/Qwen}},
  note = {Accessed: 2026-06-17},
}

@inproceedings{ragusa2024enigma51,
  title = {{ENIGMA-51}: Towards a Fine-Grained Understanding of Human Behavior in Industrial Scenarios},
  author = {Ragusa, Francesco and Leonardi, Rosario and Mazzamuto, Michele and Bonanno, Claudia and Scavo, Rosario and Furnari, Antonino and Farinella, Giovanni Maria},
  booktitle = {Proceedings of the IEEE/CVF Winter Conference on Applications of Computer Vision ({WACV})},
  pages = {7065--7075},
  year = {2024},
  doi = {10.1109/WACV57701.2024.00449},
  url = {https://openaccess.thecvf.com/content/WACV2024/html/Ragusa_ENIGMA-51_Towards_a_Fine-Grained_Understanding_of_Human_Behavior_in_Industrial_WACV_2024_paper.html},
}

@inproceedings{ramakrishnan2023naq,
  title = {{NaQ}: Leveraging narrations as queries to supervise episodic memory},
  author = {Ramakrishnan, Santhosh Kumar and Al-Halah, Ziad and Grauman, Kristen},
  booktitle = {Proceedings of the IEEE/CVF Conference on Computer Vision and Pattern Recognition ({CVPR})},
  pages = {18994--19004},
  year = {2023},
  doi = {10.1109/CVPR52729.2023.00647},
  url = {https://doi.org/10.1109/CVPR52729.2023.00647},
}

@inproceedings{rossetto2025castle,
  title = {The {CASTLE} 2024 Dataset: Advancing the Art of Multimodal Understanding},
  author = {Luca Rossetto and Werner Bailer and Duc-Tien Dang-Nguyen and Graham Healy and Björn Þór Jónsson and Onanong Kongmeesub and Hoang-Bao Le and Stevan Rudinac and Klaus Schöffmann and Florian Spiess and Allie Tran and Minh-Triet Tran and Quang-Linh Tran and Cathal Gurrin},
  booktitle = {Proceedings of the 33rd ACM International Conference on Multimedia},
  pages = {12629--12635},
  year = {2025},
  doi = {10.1145/3746027.3758199},
  url = {https://doi.org/10.1145/3746027.3758199},
}

@inproceedings{song2024moviechat,
  title = {{MovieChat}: From Dense Token to Sparse Memory for Long Video Understanding},
  author = {Enxin Song and Wenhao Chai and Guanhong Wang and Yucheng Zhang and Haoyang Zhou and Feiyang Wu and Haozhe Chi and Xun Guo and Tian Ye and Yanting Zhang and Yan Lu and Jenq-Neng Hwang and Gaoang Wang},
  booktitle = {Proceedings of the IEEE/CVF Conference on Computer Vision and Pattern Recognition ({CVPR})},
  pages = {18221--18232},
  year = {2024},
  doi = {10.1109/CVPR52733.2024.01725},
  url = {https://openaccess.thecvf.com/content/CVPR2024/html/Song_MovieChat_From_Dense_Token_to_Sparse_Memory_for_Long_Video_CVPR_2024_paper.html},
}

@inproceedings{tang2025adaptive,
  title = {Adaptive Keyframe Sampling for Long Video Understanding},
  author = {Tang, Xi and Qiu, Jihao and Xie, Lingxi and Tian, Yunjie and Jiao, Jianbin and Ye, Qixiang},
  booktitle = {Proceedings of the IEEE/CVF Conference on Computer Vision and Pattern Recognition ({CVPR})},
  pages = {29118--29128},
  year = {2025},
  doi = {10.1109/CVPR52734.2025.02711},
  url = {https://openaccess.thecvf.com/content/CVPR2025/html/Tang_Adaptive_Keyframe_Sampling_for_Long_Video_Understanding_CVPR_2025_paper.html},
}

@inproceedings{tian2025identifying,
  title = {Identifying and Mitigating Position Bias of Multi-image Vision-Language Models},
  author = {Tian, Xinyu and Zou, Shu and Yang, Zhaoyuan and Zhang, Jing},
  booktitle = {Proceedings of the IEEE/CVF Conference on Computer Vision and Pattern Recognition ({CVPR})},
  pages = {10599--10609},
  year = {2025},
  doi = {10.1109/CVPR52734.2025.00991},
  url = {https://openaccess.thecvf.com/content/CVPR2025/html/Tian_Identifying_and_Mitigating_Position_Bias_of_Multi-image_Vision-Language_Models_CVPR_2025_paper.html},
}

@inproceedings{tokmurziyev2025llmglasses,
  title = {{LLM-Glasses}: {GenAI}-Driven Glasses With Haptic Feedback for Navigation of Visually Impaired People},
  author = {Tokmurziyev, Issatay and Altamirano Cabrera, Miguel and Khan, Muhammad Haris and Mahmoud, Yara and Tsetserukou, Dzmitry},
  booktitle = {Companion Proceedings of the 21st ACM/IEEE International Conference on Human-Robot Interaction},
  pages = {994--998},
  year = {2026},
  doi = {10.1145/3776734.3794543},
  url = {https://doi.org/10.1145/3776734.3794543},
}

@incollection{tulving1972episodic,
  title = {Episodic and semantic memory},
  author = {Tulving, Endel},
  booktitle = {Organization of Memory},
  editor = {Tulving, Endel and Donaldson, Wayne},
  pages = {381--403},
  publisher = {Academic Press},
  address = {New York},
  year = {1972},
}

@article{tulving2002episodic,
  title = {Episodic Memory: From Mind to Brain},
  author = {Tulving, Endel},
  journal = {Annual Review of Psychology},
  volume = {53},
  number = {1},
  pages = {1--25},
  year = {2002},
  doi = {10.1146/annurev.psych.53.100901.135114},
  url = {https://doi.org/10.1146/annurev.psych.53.100901.135114},
}

@article{tulving1973encoding,
  title = {Encoding Specificity and Retrieval Processes in Episodic Memory},
  author = {Tulving, Endel and Thomson, Donald M.},
  journal = {Psychological Review},
  volume = {80},
  number = {5},
  pages = {352--373},
  year = {1973},
  doi = {10.1037/h0020071},
  url = {https://doi.org/10.1037/h0020071},
}

@article{wang2025internvl35,
  title = {{InternVL3.5}: Advancing Open-Source Multimodal Models in Versatility, Reasoning, and Efficiency},
  author = {Weiyun Wang and Zhangwei Gao and Lixin Gu and Hengjun Pu and Long Cui and Xingguang Wei and Zhaoyang Liu and Linglin Jing and Shenglong Ye and Jie Shao and Zhaokai Wang and Zhe Chen and Hongjie Zhang and Ganlin Yang and Haomin Wang and Qi Wei and Jinhui Yin and Wenhao Li and Erfei Cui and Guanzhou Chen and Zichen Ding and Changyao Tian and Zhenyu Wu and Jingjing Xie and Zehao Li and Bowen Yang and Yuchen Duan and Xuehui Wang and Zhi Hou and Haoran Hao and Tianyi Zhang and Songze Li and Xiangyu Zhao and Haodong Duan and Nianchen Deng and Bin Fu and Yinan He and Yi Wang and Conghui He and Botian Shi and Junjun He and Yingtong Xiong and Han Lv and Lijun Wu and Wenqi Shao and Kaipeng Zhang and Huipeng Deng and Biqing Qi and Jiaye Ge and Qipeng Guo and Wenwei Zhang and Songyang Zhang and Maosong Cao and Junyao Lin and Kexian Tang and Jianfei Gao and Haian Huang and Yuzhe Gu and Chengqi Lyu and Huanze Tang and Rui Wang and Haijun Lv and Wanli Ouyang and Limin Wang and Min Dou and Xizhou Zhu and Tong Lu and Dahua Lin and Jifeng Dai and Weijie Su and Bowen Zhou and Kai Chen and Yu Qiao and Wenhai Wang and Gen Luo},
  journal = {arXiv preprint arXiv:2508.18265},
  year = {2025},
  eprint = {2508.18265},
  archiveprefix = {arXiv},
  primaryclass = {cs.CV},
  url = {https://arxiv.org/abs/2508.18265},
}

@article{wang2026egomemreason,
  title = {{EgoMemReason}: A Memory-Driven Reasoning Benchmark for Long-Horizon Egocentric Video Understanding},
  author = {Wang, Ziyang and Zhang, Yue and Yu, Shoubin and Zhang, Ce and Zhao, Zengqi and Yoon, Jaehong and Lee, Hyunji and Bertasius, Gedas and Bansal, Mohit},
  journal = {arXiv preprint arXiv:2605.09874},
  year = {2026},
  eprint = {2605.09874},
  archiveprefix = {arXiv},
  primaryclass = {cs.CV},
  url = {https://arxiv.org/abs/2605.09874},
}

@inproceedings{xing2025caprl,
  title = {{CapRL}: Stimulating Dense Image Caption Capabilities via Reinforcement Learning},
  author = {Long Xing and Xiaoyi Dong and Yuhang Zang and Yuhang Cao and Jianze Liang and Qidong Huang and Jiaqi Wang and Feng Wu and Dahua Lin},
  booktitle = {The Fourteenth International Conference on Learning Representations},
  year = {2026},
  url = {https://openreview.net/forum?id=JLelnhqXaC},
}

@inproceedings{yang2025egolife,
  title = {{EgoLife}: Towards Egocentric Life Assistant},
  author = {Jingkang Yang and Shuai Liu and Hongming Guo and Yuhao Dong and Xiamengwei Zhang and Sicheng Zhang and Pengyun Wang and Zitang Zhou and Binzhu Xie and Ziyue Wang and Bei Ouyang and Zhengyu Lin and Marco Cominelli and Zhongang Cai and Yuanhan Zhang and Peiyuan Zhang and Fangzhou Hong and Joerg Widmer and Francesco Gringoli and Lei Yang and Bo Li and Ziwei Liu},
  booktitle = {Proceedings of the IEEE/CVF Conference on Computer Vision and Pattern Recognition ({CVPR})},
  pages = {28885--28900},
  year = {2025},
  doi = {10.1109/CVPR52734.2025.02690},
  url = {https://openaccess.thecvf.com/content/CVPR2025/html/Yang_EgoLife_Towards_Egocentric_Life_Assistant_CVPR_2025_paper.html},
}

@article{yang2025captionqa,
  title = {{CaptionQA}: Is Your Caption as Useful as the Image Itself?},
  author = {Shijia Yang and Yunong Liu and Bohan Zhai and Ximeng Sun and Zicheng Liu and Emad Barsoum and Manling Li and Chenfeng Xu},
  journal = {arXiv preprint arXiv:2511.21025},
  year = {2025},
  eprint = {2511.21025},
  archiveprefix = {arXiv},
  primaryclass = {cs.CV},
  url = {https://arxiv.org/abs/2511.21025},
}

@inproceedings{ye2025mmego,
  title = {{MM-EGO}: Towards Building Egocentric Multimodal {LLMs} for Video {QA}},
  author = {Hanrong Ye and Haotian Zhang and Erik Daxberger and Lin Chen and Zongyu Lin and Yanghao Li and Bowen Zhang and Haoxuan You and Dan Xu and Zhe Gan and Jiasen Lu and Yinfei Yang},
  booktitle = {The Thirteenth International Conference on Learning Representations},
  year = {2025},
  url = {https://openreview.net/forum?id=67sSPPAZiG},
}

@article{silvr2025,
  title = {{SiLVR}: A Simple Language-Based Video Reasoning Framework},
  author = {Zhang, Ce and Lin, Yan-Bo and Wang, Ziyang and Bansal, Mohit and Bertasius, Gedas},
  journal = {Transactions on Machine Learning Research},
  year = {2026},
  url = {https://openreview.net/forum?id=mQZbh9Zlbw},
}

@article{zhang2025llavavideo,
  title = {{LLaVA-Video}: Video Instruction Tuning With Synthetic Data},
  author = {Zhang, Yuanhan and Wu, Jinming and Li, Wei and Li, Bo and Ma, Zejun and Liu, Ziwei and Li, Chunyuan},
  journal = {Transactions on Machine Learning Research},
  year = {2025},
  url = {https://openreview.net/forum?id=EElFGvt39K},
}

@article{zhang2026watch,
  title = {Watch Before You Answer: Learning From Visually Grounded Post-Training},
  author = {Yuxuan Zhang and EunJeong Hwang and Huaisong Zhang and Penghui Du and Yiming Jia and Dongfu Jiang and Xuan He and Shenhui Zhang and Ping Nie and Peter West and Kelsey R. Allen},
  journal = {arXiv preprint arXiv:2604.05117},
  year = {2026},
  eprint = {2604.05117},
  archiveprefix = {arXiv},
  primaryclass = {cs.CV},
  url = {https://arxiv.org/abs/2604.05117},
}

\clearpage
\appendix

\section{An Introduction to Episodic Memory}
\label{appendixA}

The notion of episodic memory originates from cognitive psychology. Tulving introduced episodic memory as memory for personally experienced events anchored in specific temporal and spatial contexts, contrasting it with semantic memory, which captures general world knowledge independent of a particular episode~\cite{tulving1972episodic,tulving2002episodic}. Later theories further clarified that episodic memory is not simply a passive storage system but a constructive retrieval process. For instance, the encoding specificity principle argues that successful recall depends on whether retrieval cues overlap with the original encoded experience~\cite{tulving1973encoding}. Similarly, autobiographical memory research views human memories as a hierarchical structure, organized across lifetime periods, general events, and event-specific details~\cite{conway2000construction}.

When bridged to multi-modal artificial intelligence, these cognitive concepts translate naturally into the engineering challenges of long egocentric video understanding. In this domain, a wearable assistant is rarely asked only to retrieve semantic knowledge—such as recognizing whether a generic cup or dog appears in a frame. Instead, it must act as an externalized episodic memory, answering situated questions about the user's own experience: where a specific object was left, what action happened before another event, or which person interacted with the camera wearer. Because a continuous video stream contains dense, unrefined low-level visual observations, useful machine recall cannot rely on unstructured scanning. For example, the query ``Where did I leave my keys?'' may be answered only if the system can use object identity, location, action history, and temporal cues to retrieve a small span from a long, high-bandwidth visual record. Thus, establishing episodic memory in video understanding is fundamentally a problem of structured encoding (video-to-text translation), cue-driven retrieval (textual localization), and evidence-grounded answer generation (fine-grained visual verification).

\paragraph{Wearable memory and lifelogging.}
The connection between episodic memory and wearable devices predates modern VLMs. Early lifelogging systems such as MyLifeBits aimed to create a searchable digital archive of a person's documents, photos, audio, and daily activities~\cite{gemmell2006mylifebits,bell2009total}. Wearable camera systems such as SenseCam ~\cite{hodges2006sensecam} and later lifelogging devices explored the use of first-person visual records as memory aids, particularly for supporting autobiographical recall in users with memory impairment.~\cite{doherty2012sensecam}. These systems revealed both the promise and the core difficulty of externalized memory: recording is increasingly easy, but retrieval is hard. A lifetime archive is valuable only if the user can recover the right episode at the right time.

Modern AR glasses and wearable sensing platforms revive this question at a larger scale. Project Aria, for example, frames future AR devices as always-available, context-aware systems that collect egocentric multimodal streams for personalized AI applications~\cite{engel2023projectaria}. Recent wearable assistant applications, from navigation for visually impaired users to laboratory procedure assistance, also depend on persistent contextual memory rather than isolated frame recognition~\cite{tokmurziyev2025llmglasses,cong2025labos}. In such settings, episodic memory is not an auxiliary capability; it is the mechanism that allows an assistant to connect present queries to past observations. Without episodic memory, the device can perceive the current scene but cannot answer what happened before, where an object was last seen, or how the user's current situation relates to previous steps.

\paragraph{Episodic memory in egocentric video benchmarks.}
Egocentric video research has gradually moved from short action recognition toward memory-centered understanding. Earlier egocentric datasets such as EPIC-KITCHENS emphasized fine-grained hand-object interactions and daily activities~\cite{damen2018scaling,Damen2020RESCALING}. Ego4D marked a major shift by collecting thousands of hours of first-person video across diverse daily scenarios. Crucially, it was the first to formally introduce the cognitive concept of episodic memory into the computer vision community, explicitly defining benchmark tasks around past, present, and future understanding~\cite{grauman2022ego4d}. Its episodic memory tasks, including natural language queries, visual queries, and moment queries, formalize the problem of locating or retrieving information from a camera wearer's past experience. This design established egocentric episodic memory as a central challenge for first-person perception.

Subsequent work extended this direction in several ways. QAEGO4D, framed episodic-memory-based question answering on egocentric videos and highlighted the need for systems that can answer user-centric questions over long visual experience~\cite{Baermann_2022_CVPR}. Episodic Memory Question Answering further studied memory representations for embodied agents and AR assistants in indoor environments, requiring models to ground answers in spatio-temporal scene memory~\cite{datta2022episodic}. NaQ showed that narrations can be repurposed as supervision for natural-language episodic memory queries, improving query localization in Ego4D~\cite{ramakrishnan2023naq}. EgoSchema then pushed video-language understanding toward longer egocentric clips and diagnostic multiple-choice QA~\cite{mangalam2024egoschema}. More recently, Ego-Exo4D introduced synchronized first- and third-person recordings of skilled activities, enabling research on memory and reasoning across viewpoints~\cite{grauman2024egoexo4d}, while EgoMemReason studies memory-driven reasoning over ultra-long egocentric videos~\cite{wang2026egomemreason}.

Together, these works show that episodic memory has become a defining problem for egocentric AI. However, most existing benchmarks either focus on temporal localization, direct VideoQA, or specialized reasoning settings. They do not directly ask whether a compact textual memory, generated from the video itself, can replace raw visual input for long-horizon QA. This question is important because wearable devices face severe constraints: long egocentric streams are expensive to store, transmit, and process as visual tokens, while downstream users need fast and interpretable access to past events.

\paragraph{Why caption-based episodic memory matters.}
Caption-based memory offers a natural bridge between psychological theories of retrieval and practical constraints of long video understanding. In cognitive terms, captions act as retrieval cues: they preserve discrete descriptions of objects, actions, locations, and timestamps that can later be matched to a user's question. In systems terms, captions compress high-bandwidth visual streams into searchable textual logs, reducing the cost of querying hours of video and mitigating the long-context decay caused by severe visual-token overload. This is especially attractive for wearable assistants, where the device may need to support many downstream tasks, such as object finding, safety reminders, procedural guidance, navigation, health monitoring, and human-robot collaboration.

At the same time, captions are lossy. They may omit subtle attributes, spatial relations, transient gestures, or small objects that are visually present but not selected by the captioner. This creates a central trade-off for episodic memory systems: textual memory improves scalability and accessibility, but raw visual evidence remains necessary for fine-grained verification. This trade-off motivates our benchmark design. CapMem evaluates whether self-generated timestamped captions can serve as episodic memory for long egocentric QA, while retaining evidence timestamps and fine-grained visual tags to diagnose when captions succeed or fail. The retrieve-and-verify harness further reflects a practical memory strategy: use captions as a semantic index to identify likely moments, then selectively revisit sparse visual evidence when details matter. In this sense, CapMem connects the cognitive view of episodic memory as cue-driven recall with the engineering need for scalable, evidence-grounded wearable intelligence.

\section{Guiding Taxonomy and Question Template for QA Generation} \label{temp}
A central design goal of CapMem is to evaluate the capability of VLMs to utilize textual captions as episodic memory for long egocentric videos in a way that is both comprehensive and diagnostically meaningful. To this end, to systematically assist human annotators in brainstorming and constructing high-quality QA pairs, we build our question generation pipeline around a guiding taxonomy with two orthogonal axes: \textbf{subject of inquiry} and \textbf{query intent}. The first axis distinguishes whether a question concerns the \textbf{camera wearer} and their actions, states, or attributes (\textbf{Ego}), or instead targets \textbf{external entities} visible in the scene (\textbf{Exo}). This distinction is important because egocentric memory naturally involves both self-centered and scene-centered recall. Many real-world assistant queries are inherently ego-centric, such as what the wearer picked up, where they placed an object, or what action they performed before a later event. At the same time, long-form first-person videos also contain rich exo-centric information about other people, objects, animals, tools, and environmental changes. Separating Ego from Exo therefore prevents the benchmark from collapsing into a purely self-action dataset and enables analysis of whether models fail differently when tracking the wearer versus the surrounding world.

The second axis organizes questions by \textbf{4W1H intent}: \textbf{Who}, \textbf{What}, \textbf{Where}, \textbf{When}, and \textbf{How}. This axis is motivated by the observation that episodic memory queries are not homogeneous. Some questions ask for identity (Who), others for object or event content (What), spatial placement (Where), temporal order or timing (When), or process, state transition, and manner (How). These forms correspond to different retrieval demands. For example, ``Where'' questions emphasize spatial grounding, ``When'' questions emphasize temporal localization and sequence memory, and ``How'' questions often require integrating multiple observations into a coherent procedural or relational interpretation. By combining the Ego/Exo and 4W1H axes, the taxonomy provides human annotators with a structured cognitive space of question types that spans both \emph{who or what is being queried} and \emph{what kind of recall operation is required}. This design makes the benchmark more interpretable than a flat collection of questions, because errors can be attributed to specific memory dimensions rather than only to overall accuracy.

To operationalize this taxonomy for annotators, we design
50 standardized question templates, listed in
Table~\ref{tab:50_templates}, as scaffolds for manual annotation. The templates are not intended to rigidly constrain linguistic form; instead, they provide controlled structural patterns that encourage broad coverage of episodic reasoning behaviors. In practice, they capture recurring forms of memory queries that arise in long egocentric videos, including object counting, state changes, action destinations, ego--exo interactions, attribute comparison, spatial relations, and temporally ordered events. This template-based design provides several benefits. First, it improves coverage: annotators are explicitly guided to consider diverse memory phenomena rather than overproducing only the most obvious object or action questions. Second, it improves consistency: questions across different scenarios follow comparable structural principles, making benchmark difficulty less dependent on arbitrary annotator style. Third, it improves diagnostic value: because each template is tied to a known query form, downstream evaluation can more clearly relate performance differences to specific episodic reasoning skills.

The use of templates is particularly important for CapMem because our goal is not merely to test whether a model can answer generic video questions, but whether it can recover \emph{fine-grained and time-specific} details from long first-person experience. Open-ended human annotation without structural guidance often drifts toward either overly easy questions, which can be solved from commonsense or coarse scene priors, or overly idiosyncratic questions, which are difficult to compare across videos. The 50-template design helps strike a balance between naturalness and control. Human annotators still instantiate each template with concrete entities, actions, and events from the source video, preserving realism and grounding, but the template skeleton ensures that the resulting dataset remains systematically organized. Because the questions are generated under explicit structural guidance, they can be paired with downstream filtering to remove linguistic shortcuts and ambiguous cases. Moreover, the taxonomy interacts naturally with our fine-grained visual tags, evidence timestamps, and evaluation dimensions. They define what kinds of episodic memory behaviors are tested, how those behaviors are distributed, and how model failures can later be interpreted.

\begin{table*}[t]
\centering
\scriptsize
\setlength{\tabcolsep}{3pt}
\renewcommand{\arraystretch}{0.95}
\caption{The 50 foundational question templates organized by 4W1H intent and Ego/Exo perspective.}
\label{tab:50_templates}
\begin{tabular}{c c c p{0.70\textwidth}}
\toprule
\textbf{ID} & \textbf{Intent} & \textbf{Perspective} & \textbf{Question Template} \\
\midrule
1 & How & Exo & How many \{object\_x\_plural\} were there when \{person\_z\} \{action\} \{object\_y\}? \\
2 & How & Exo & How many \{object\_x\_plural\} are at \{place\_y\}? \\
3 & How & Exo & How many \{object\_x\_plural\} appear during \{time\_interval\_y\}? \\
4 & How & Exo & How many \{object\_x\_plural\} were around when \{object\_y\} \{state\_or\_action\}? \\
5 & How & Ego & How long did I spend doing \{activity\_x\}? \\
6 & How & Exo & How long did \{someone\_x\} spend doing \{activity\_x\}? \\
7 & How & Ego & How did I \{action\} \{object\_y\}? \\
\midrule
8 & Where & Exo & Where was \{object\_x\} when I \{action\} with \{person\_y\}? \\
9 & Where & Ego & Where was I when I \{action\} with \{person\_y\}? \\
10 & Where & Exo & Where was \{object\_x\} when I \{action\} with \{object\_y\}? \\
11 & Where & Exo & Where was \{object\_x\} when \{someone\_x\} \{action\} \{object\_y\}? \\
12 & Where & Ego & Where was I when I \{action\} \{object\_y\}? \\
13 & Where & Ego & Where did I spend most of my time during \{interval\_or\_activity\_x\}? \\
14 & Where & Exo & Where did \{someone\_x\} spend most of their time during \{interval\_or\_activity\_x\}? \\
15 & Where & Ego & Where did \{object\_x\} end up after I \{action\} it? \\
\midrule
16 & When & Ego & When did I \{action\} with \{person\_y\}? \\
17 & When & Ego & When did I \{action\} at \{place\_y\}? \\
18 & When & Ego & When did I \{action\} with \{object\_y\}? \\
19 & When & Exo & When did \{someone\_x\} \{action\} with \{person\_y\}? \\
20 & When & Exo & When did \{someone\_x\} \{action\} at \{place\_y\}? \\
21 & When & Exo & When did \{someone\_x\} \{action\} with \{object\_y\}? \\
22 & When & Exo & When did \{object\_x\} become \{state\_y\}? \\
\midrule
23 & What & Ego & What did I do when I was with \{person\_y\}? \\
24 & What & Ego & What did I do at \{place\_y\}? \\
25 & What & Ego & What did I do with \{object\_y\}? \\
26 & What & Ego & What did I do during \{interval\_x\}? \\
27 & What & Exo & What did \{someone\_x\} do when \{person\_z\} \{action\}? \\
28 & What & Exo & What did \{someone\_x\} do when \{I/someone\} \{action\} at \{place\_y\}? \\
29 & What & Exo & What did \{someone\_x\} do with \{object\_y\}? \\
30 & What & Exo & What did \{someone\_x\} do during \{interval\_x\}? \\
31 & What & Exo & What \{object\_x\} did I \{action\}? \\
32 & What & Exo & Which \{object\_x\} is \{state\_y\}? \\
33 & What & Exo & What is the state of \{object\_x\}? \\
34 & What & Exo & What is the relation between \{object\_x\} and \{object\_y\}? \\
35 & What & Exo & What (object) did I use for \{activity\_x\}? \\
36 & What & Exo & What object did I use/hold when \{action\} with \{person\_x\}? \\
37 & What & Exo & What color is \{object\_x\}? \\
38 & What & Exo & What text/symbol is visible on \{object\_x\}? \\
39 & What & Ego & Which hand did I use to \{action\} with \{object\_x\}? \\
40 & What & Ego & What did I pick up from \{place\_y\}? \\
\midrule
41 & Who & Ego & Who did I \{action\} at \{place\_y\}? \\
42 & Who & Ego & Who did I \{action\} \{object\_y\} with? \\
43 & Who & Ego & Who \{action\} with me when I \{action\} with \{person\_y\}? \\
44 & Who & Ego & Who did I \{action\} with during \{interval\_x\}? \\
45 & Who & Exo & Who did \{someone\_x\} \{action\} at \{place\_y\}? \\
46 & Who & Exo & Who did \{someone\_x\} \{action\} \{object\_y\} with? \\
47 & Who & Exo & Who \{action\} \{someone\_x\} when \{someone\_z\} \{action\} with \{person\_y\}? \\
48 & Who & Ego & Who was \{person\_y\} to me during/at \{event\_or\_time\_x\}? \\
49 & Who & Ego & Who was looking at me when I \{action\}? \\
50 & Who & Exo & Who was holding/using \{object\_x\} at \{place\_y\}? \\
\bottomrule
\end{tabular}
\end{table*}

\section{Prompts} \label{prompt}
\begin{figure*}[t]
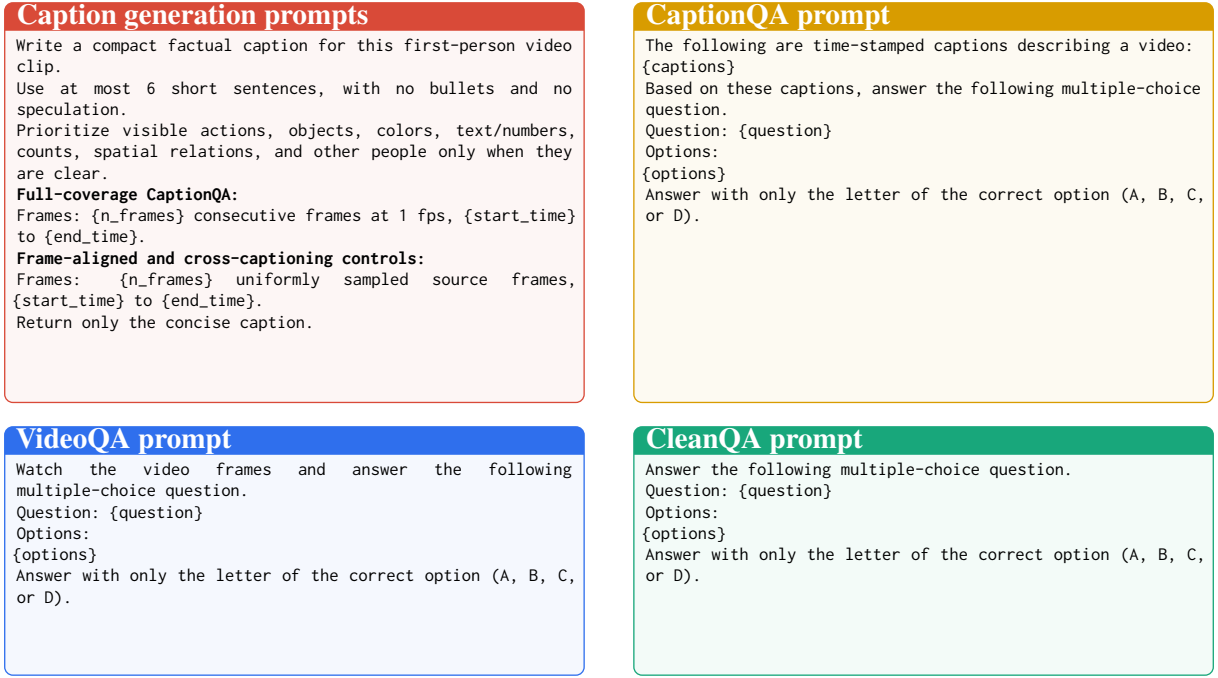

\centering

\begin{minipage}[t]{0.48\textwidth}
\vspace{0pt}
\begin{evalpromptbox}
    {PromptRed}
    {Caption generation prompts}
    {5.3cm}
Write a compact factual caption for this first-person video clip.

Use at most 6 short sentences, with no bullets and no speculation.

Prioritize visible actions, objects, colors, text/numbers, counts,
spatial relations, and other people only when they are clear.

\textbf{Full-coverage CaptionQA:}

Frames: \{n\_frames\} consecutive frames at 1 fps,
\{start\_time\} to \{end\_time\}.

\textbf{Frame-aligned and cross-captioning controls:}

Frames: \{n\_frames\} uniformly sampled source frames,
\{start\_time\} to \{end\_time\}.

Return only the concise caption.
\end{evalpromptbox}
\end{minipage}
\hfill
\begin{minipage}[t]{0.48\textwidth}
\vspace{0pt}
\begin{evalpromptbox}
    {PromptYellow}
    {CaptionQA prompt}
    {5.3cm}
The following are time-stamped captions describing a video:

\{captions\}

Based on these captions, answer the following multiple-choice question.

Question: \{question\}

Options:

\{options\}

Answer with only the letter of the correct option (A, B, C, or D).
\end{evalpromptbox}
\end{minipage}

\par\vspace{0.7em}

\begin{minipage}[t]{0.48\textwidth}
\vspace{0pt}
\begin{evalpromptbox}
    {PromptBlue}
    {VideoQA prompt}
    {3.3cm}
Watch the video frames and answer the following multiple-choice question.

Question: \{question\}

Options:

\{options\}

Answer with only the letter of the correct option (A, B, C, or D).
\end{evalpromptbox}
\end{minipage}
\hfill
\begin{minipage}[t]{0.48\textwidth}
\vspace{0pt}
\begin{evalpromptbox}
    {PromptGreen}
    {CleanQA prompt}
    {3.3cm}
Answer the following multiple-choice question.

Question: \{question\}

Options:

\{options\}

Answer with only the letter of the correct option (A, B, C, or D).
\end{evalpromptbox}
\end{minipage}

\caption{Evaluation prompts used for caption generation, CaptionQA,
VideoQA, and CleanQA. Caption generation uses shared instructions with
setting-specific frame descriptions: consecutive 1-FPS frames for
full-coverage CaptionQA and uniformly sampled source frames for the
frame-aligned and cross-captioning controls.}
\label{fig:evaluation_prompts}
\vspace{-6pt}
\end{figure*}

To ensure a fair comparison across model families, we use the same task-specific prompts for all evaluated models whenever the input setting is the same. In particular, the caption generation prompt is shared across open-source and closed-source models, and the same CaptionQA, VideoQA, and CleanQA prompts are used for their corresponding evaluation settings. The only differences across the 30s and 60s CaptionQA settings are the caption window length and the resulting caption content; the prompt template itself remains unchanged. The propmts can be found in Figure 
\ref{fig:evaluation_prompts}.


\section{Harness Engineering}\label{harness}

\begin{figure}
    \centering
    \vspace{-10pt}
    \includegraphics[
        width=\linewidth,
        trim=0 10 0 10,
        clip
    ]{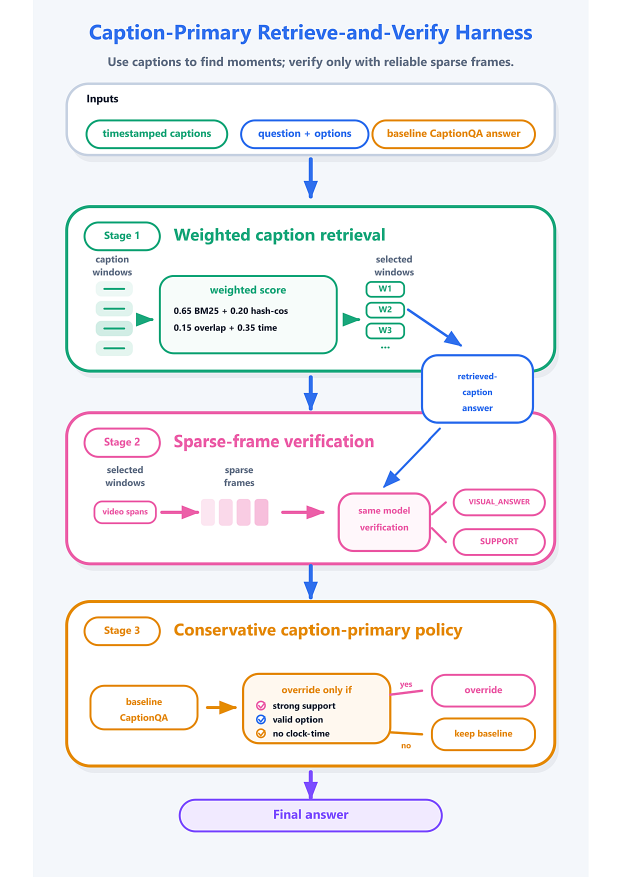}
    \caption{Caption-primary retrieve-and-verify harness.}
    \label{fig:harness_flowchart}
\end{figure}

We implement a caption-primary retrieve-and-verify harness to test whether sparse visual evidence can repair errors made by caption-based long-video reasoning. As summarized in
Figure~\ref{fig:harness_flowchart}, the harness is intentionally conservative: the baseline CaptionQA prediction remains the default answer, and visual verification is only allowed to override it under strong evidence.

\paragraph{Inputs.}
For each evaluated question, the harness uses three inputs: the question with four answer options, cached timestamped captions generated under a fixed caption specification, and the initial CaptionQA baseline prediction (i.e., the original answer derived exclusively from the text-based reasoning stage before any visual verification). We do not mix caption granularities: results using 30s captions are compared only against the corresponding 30s CaptionQA baseline, and likewise for 60s captions.

\paragraph{Caption-window retrieval.}
For each multiple-choice question, we form a retrieval query by concatenating the question and all answer options. The harness then scores each timestamped caption window using a deterministic weighted retrieval rule:
\[
\begin{aligned}
s ={}& 0.65 \cdot s_{\text{BM25}}
    + 0.20 \cdot s_{\text{hash-cos}} \\
   &+ 0.15 \cdot s_{\text{overlap}}
    + 0.35 \cdot s_{\text{time}} .
\end{aligned}
\]
We decouple the retrieval score into four distinct components because relying on a single  metric is brittle when querying self-generated captions. Each term targets a different failure mode of textual retrieval:
\begin{itemize}\item \textbf{BM25 ($s_{\text{BM25}}$)} acts as the primary retrieval backbone. By leveraging term frequency and inverse document frequency (TF-IDF), it heavily rewards exact matches of rare, discriminative keywords (e.g., a specific object name) while discounting ubiquitous stop words.\item \textbf{Hash-Cosine ($s_{\text{hash-cos}}$)} provides a lightweight, fuzzy similarity signal based on n-gram hashing. Unlike the strict exact-match requirement of BM25, this term improves robustness against slight phrasing variations, synonyms, or morphological differences between the question and the generated caption.\item \textbf{Word Overlap ($s_{\text{overlap}}$)} computes the unweighted density of shared content words. This serves as a critical counterbalance to BM25: it prevents the system from over-indexing on a single extremely rare word while ignoring the rest of the query, ensuring that the retrieved window broadly covers the multiple entities mentioned in the prompt.\item \textbf{Time-Hint ($s_{\text{time}}$)} acts as a temporal grounding mechanism. Pure lexical models are notoriously blind to temporal logic. This term explicitly boosts windows that match temporal markers (e.g., before'', after'', first'', last'') or explicit timestamps present in the query, which is essential for episodic memory tasks involving sequence and state changes.\end{itemize}
Together, these four signals form a robust ensemble that accurately localizes relevant moments without the computational overhead of deploying a neural bi-encoder for every frame.

We use the following retrieval configuration:
\[
\begin{aligned}
&\texttt{top\_k} = 3,\\
&\texttt{nms\_seconds} = 45,\\
&\texttt{neighbor} = 1.
\end{aligned}
\]
Here, \texttt{top\_k}=3 means that the retriever keeps up to three highest-scoring seed caption windows for each question. To avoid selecting near-duplicate temporal evidence, \texttt{nms\_seconds}=45 suppresses additional seed windows that fall within 45 seconds of an already selected seed. Finally, \texttt{neighbor}=1 expands each selected seed by adding one adjacent caption window before and after it, when available, as supporting evidence. Thus, the retrieved evidence contains at most nine caption windows: up to three temporally separated seed windows, each expanded with one neighboring window on both sides. Given the caption cache, question/options, and retrieval configuration, this evidence set is fixed and does not depend on model-generated timestamps. The length of each caption window follows the caption cache granularity: 30 seconds for the 30s setting and 60 seconds for the 60s setting, except for boundary windows such as the final segment.

\paragraph{Retrieved-caption answering.}
After retrieval, the evaluated model answers the question based only on the caption text associated with the retrieved video windows. It outputs a retrieved-caption answer and a confidence label:
\[
\begin{aligned}
&\texttt{ANSWER} \in \{A,B,C,D\},\\
&\texttt{CONFIDENCE} \in \{\text{low},\text{medium},\text{high}\}.
\end{aligned}
\]
This answer is not used as the final prediction. Instead, it serves as an intermediate candidate answer that focuses the subsequent visual verification step.
\paragraph{Sparse-frame verification.}
The harness then constructs short visual clips from the same retrieved windows. We use:
\[
\begin{aligned}
&\texttt{max\_clips} = 3,\\
&\texttt{clip\_neighbor} = 1,\\
&\texttt{expand\_sec} = 5,\\
&\texttt{frames\_per\_clip} = 4.
\end{aligned}
\]
Here, \texttt{max\_clips}=3 limits verification to at most three visual intervals. Although the retrieved caption evidence may contain up to nine caption windows after neighbor expansion, the visual verification stage is centered only on the retrieved seed windows. Since \texttt{top\_k}=3 selects at most three seed windows, \texttt{max\_clips}=3 yields at most three visual intervals. For each visual interval, \texttt{clip\_neighbor}=1 merges one adjacent caption window before and after the seed window when available. The merged window is then expanded by \texttt{expand\_sec}=5 seconds on both sides and clipped to the video boundaries. Finally, \texttt{frames\_per\_clip}=4 sparse frames are uniformly sampled from each expanded interval. Thus, each question uses at most \(3 \times 4 = 12\) frames for visual verification. The same evaluated model is then called again with a verification-specific prompt. This prompt provides the sparse frames, the question/options, and the retrieved-caption answer as context for judging the visual evidence. The goal is to determine which option is supported by the sampled frames and how strong the support is, rather than asking the model to solve the full long-video task from scratch. Output of this step is:
\[
\begin{aligned}
&\texttt{VISUAL\_ANSWER} \in \{A,B,C,D,\text{UNKNOWN}\},\\
&\texttt{VISUAL\_SUPPORT} \in \{\text{strong},\text{weak},\text{none}\}.
\end{aligned}
\]

\paragraph{Conservative caption-primary policy.}
The final decision follows a conservative policy named \texttt{visual\_strong\_notime}. Let \(a_{\text{base}}\) be the  initial CaptionQA 
baseline prediction. By default, the final answer is:
\[
a_{\text{final}} = a_{\text{base}}.
\]
The harness overrides this baseline only when all of the following conditions hold: the visual support is \texttt{strong}, the visual answer is an option in \(\{A,B,C,D\}\), and the question/options do not contain an explicit clock-time cue such as \texttt{01:23} or \texttt{01:23:45}. Formally,
\[
\begin{aligned}
a_{\text{final}} =
\begin{cases}
a_{\text{visual}},
& \begin{aligned}[t]
  &\text{if support is strong},\\
  &a_{\text{visual}}\in\{A,B,C,D\},\\
  &\text{and no clock-time cue exists},
  \end{aligned}\\
a_{\text{base}},
& \text{otherwise}.
\end{cases}
\end{aligned}
\]
We block overrides for explicit clock-time questions because sparse frames may miss the referenced moment.

CapMem currently contains a single 1,000-question evaluation split and does not provide a separate development set. The retrieval weights, \(top\_k=3\), \(nms\_seconds=45\), and visual-verification budget were set as heuristic defaults rather than selected through an accuracy-based search. However, the final override policy, including the clock-time rule, was chosen after inspecting errors and comparing policy variants on the full benchmark. The resulting retrieve-and-verify gains should therefore be interpreted as a post-hoc proof of concept rather than an unbiased estimate of out-of-sample improvement.
\paragraph{Reproducibility.}
The released script takes the CaptionQA 
baseline prediction result JSON, the raw retrieval-frame harness JSON, and the question file as inputs. It reports the final accuracy, the  CaptionQA baseline accuracy, the accuracy delta, the number of overrides, and whether each override changes a baseline-wrong example into a correct final prediction or a baseline-correct example into an incorrect final prediction. This separation keeps the final evaluation auditable: retrieval and sparse-frame verification provide candidate correction signals, while the final answer remains caption-primary unless the visual evidence is strong.

\section{Annotation and Quality-Control Pipeline}
\label{app:annotation_pipeline}

CapMem prioritizes manually verified questions rather than automatic
benchmark expansion. Table~\ref{tab:annotation_funnel} summarizes
the complete construction pipeline. Twelve trained annotators first
created 2,109 candidate QA pairs using the taxonomy and templates
described in Appendix ~\ref{temp}. Two reviewers
then checked the solvability, answers and template and category assignments, resolved
disagreements by consensus, and retained 1,276 candidates.

A four-model text-only ensemble subsequently identified 352 questions
that could be answered from the question and options without video
access, leaving 924 candidates. Because this automated filter could
also reject visually grounded questions, the reviewers re-examined
all rejected candidates and restored 76 false positives. They finally
verified that each of the resulting 1,000 questions was unambiguously
answerable from the source video.

\begin{table*}[t]
\centering
\small
\setlength{\tabcolsep}{5pt}
\renewcommand{\arraystretch}{1.08}
\begin{tabularx}{\textwidth}{
    l
    c
    >{\raggedright\arraybackslash}X
    c
}
\toprule
Stage & Input & Procedure & Output \\
\midrule
Candidate creation
& -- 
& Twelve trained annotators manually constructed QA pairs.
& 2,109 \\

Internal review
& 2,109
& Two reviewers checked answers and template/category assignments
and resolved disagreements by consensus.
& 1,276 \\

Text-only filtering
& 1,276
& A four-model ensemble removed questions answerable without
video access.
& 924 \\

Manual adjudication and visual verification
& \makecell{924 retained\\+ 352 rejected}
& Reviewers re-examined the automatically rejected candidates,
restored 76 false positives, and verified visual solvability.
& 1,000 \\
\bottomrule
\end{tabularx}
\caption{Annotation and quality-control pipeline. The final benchmark
retains 1,000 of 2,109 candidate questions.}
\label{tab:annotation_funnel}
\end{table*}

Fine-grained visual tags were assigned according to explicit definitions and checked during review. Specifically, Object denotes questions requiring identification, recognition, or counting of visible entities; Attribute covers visually observable properties such as color, shape, state, text, or number; Spatial captures location, direction, containment, and relative spatial relations; and Action denotes visible actions, interactions, or state transitions. Tags are multi-label and are assigned according to the fine-grained visual information explicitly queried by each question. Reviewer disagreements were resolved by consensus.

The final benchmark covers 33.7 hours, 16 scenarios, and four
application domains. Its scale reflects the cost of manual
construction, review, shortcut filtering, and visual-solvability
verification. It should therefore be interpreted as a
quality-controlled evaluation set rather than a comprehensive sample
of all wearable-video settings. Release artifacts are detailed in
Appendix~\ref{app:licensing}.

\section{Supplementary Experimental Settings and Results}\label{aE}
This appendix reports the model, deployment, decoding, and API
configurations required to reproduce the experiments. It then provides
video-clustered bootstrap uncertainty estimates for the main paired
comparisons, complete fine-grained and evidence-location results,
examines whether the observed duration crossover is directly caused by
the VideoQA frame cap, and reports measured frame, runtime, token, and
cost accounting.

\subsection{Supplementary Experimental Settings}\label{app:evaluation_settings}

\paragraph{Evaluation pipeline.}
All visual inputs are drawn from a 1-FPS frame cache. In full-coverage
CaptionQA, consecutive frames are grouped into non-overlapping
30- or 60-second windows, and each window is captioned once. The
timestamped captions are then reused by every question associated
with the video; no visual frames are accessed during downstream QA.
Direct VideoQA instead receives deterministically and uniformly
sampled frames for each question. In the frame-aligned control,
CaptionQA uses exactly the same selected frames as Direct VideoQA,
with all other model and decoding settings unchanged.

\begin{table*}
\centering
\scriptsize
\setlength{\tabcolsep}{5pt}
\renewcommand{\arraystretch}{1.06}
\begin{tabularx}{\textwidth}{
    l
    >{\raggedright\arraybackslash}X
    >{\raggedright\arraybackslash}X
}
\toprule
Model & Hugging Face model ID & Exact revision \\
\midrule
InternVL3.5-8B
& \path{OpenGVLab/InternVL3_5-8B}
& \path{9bb6a56ad9cc69db95e2d4eeb15a52bbcac4ef79} \\

InternVL3.5-38B
& \path{OpenGVLab/InternVL3_5-38B}
& \path{de99855be3642cd44fe97c9b72d70e5ce2c07f69} \\

Qwen3-VL-2B
& \path{Qwen/Qwen3-VL-2B-Instruct}
& \path{89644892e4d85e24eaac8bacfd4f463576704203} \\

Qwen3-VL-4B
& \path{Qwen/Qwen3-VL-4B-Instruct}
& \path{ebb281ec70b05090aa6165b016eac8ec08e71b17} \\

Qwen3-VL-8B
& \path{Qwen/Qwen3-VL-8B-Instruct}
& \path{0c351dd01ed87e9c1b53cbc748cba10e6187ff3b} \\

Qwen3-VL-32B
& \path{Qwen/Qwen3-VL-32B-Instruct}
& \path{0cfaf48183f594c314753d30a4c4974bc75f3ccb} \\

Qwen3.5-2B
& \path{Qwen/Qwen3.5-2B}
& \path{15852e8c16360a2fea060d615a32b45270f8a8fc} \\

Qwen3.5-4B
& \path{Qwen/Qwen3.5-4B}
& \path{851bf6e806efd8d0a36b00ddf55e13ccb7b8cd0a} \\

Qwen3.5-9B
& \path{Qwen/Qwen3.5-9B}
& \path{c202236235762e1c871ad0ccb60c8ee5ba337b9a} \\

Qwen3.5-27B
& \path{Qwen/Qwen3.5-27B}
& \path{b7ca741b86de18df552fd2cc952861e04621a4bd} \\
\bottomrule
\end{tabularx}
\caption{Exact checkpoints used for the open-weight model evaluations. The checkpoints were retrieved on February 15, 2026 for
InternVL3.5 and Qwen3-VL, March 4, 2026 for Qwen3.5-27B, and
March 11, 2026 for the remaining Qwen3.5 models (UTC).}
\label{tab:open_checkpoint_revisions}
\end{table*}
\paragraph{Open-weight models.}
Table~\ref{tab:open_checkpoint_revisions} lists the exact Hugging
Face checkpoint identifiers and revisions, while
Table~\ref{tab:open_runtime_configuration} summarizes the hardware,
parallelism, context lengths, frame caps, and maximum observed input
lengths. Open-weight inference uses vLLM 0.19.0 and
Transformers 4.57.6 with Python 3.10.20, PyTorch 2.10.0,
CUDA 12.8, and cuDNN 9.10.2.21. Models run in BF16 without
quantization on NVIDIA H200 GPUs. InternVL3.5-8B uses one GPU,
InternVL3.5-38B uses two-way tensor parallelism, and all Qwen
models use one GPU. No data or pipeline parallelism is used.
Prefix caching is enabled.

Caption generation batches eight windows per
\texttt{llm.generate} call. CaptionQA submits the 1,000 questions
in one application-level call, while VideoQA groups questions by
video. Caption generation, CaptionQA, and VideoQA use output limits
of 220, 200, and 200 tokens, respectively. All use greedy decoding
with temperature 0, $\texttt{top\_p}=1.0$, and
$\texttt{top\_k}=0$; Qwen3.5 thinking is disabled. Every observed
input remains below the corresponding context limit, and no
application-level input truncation is applied.

\paragraph{Closed-source models.}
Table~\ref{tab:closed_api_configuration} summarizes the retained
OpenRouter request parameters, retry behavior, and scoring policy.
The main GPT-5.2 and Gemini 3 Flash experiments use OpenRouter
through the \texttt{/api/v1/chat/completions} endpoint. The requested
aliases resolve to \texttt{openai/gpt-5.2-20251211} and
\texttt{google/gemini-3-flash-preview-20251217}, respectively, on
May 22, 2026. Requests use temperature 0 and minimal reasoning
effort. No candidate count, decoding seed, stop sequence, explicit
thinking budget, JSON schema, or custom safety setting is specified.
Only the first returned candidate is consumed, and provider-default
safety settings are retained. Responses are returned as plain text
and parsed into one of $\{A,B,C,D\}$.

The caption output limit is 220 tokens. The retained results indicate
CaptionQA limits of 512 tokens for GPT-5.2 and 64 tokens for
Gemini 3 Flash. Because the original complete launch commands were
not preserved, these two values are reported as recovered
configurations.

\paragraph{Retry and scoring policy.}
Closed-source requests use a 180-second timeout and up to nine total
attempts. HTTP 408, 409, 429, 500, 502, 503, and 504 responses, as
well as network exceptions, are retried with exponential backoff and
random jitter. A successful but empty, refused, safety-blocked, or
unparsable QA response is not semantically re-requested and is
counted as incorrect. Empty captions are retained, and downstream QA
proceeds without a semantic retry. Raw outputs, parsed predictions,
correctness, token counts, and finish reasons are stored for auditing.

\begin{table*}[t]
\centering
\small
\setlength{\tabcolsep}{6pt}
\renewcommand{\arraystretch}{1.06}
\begin{tabular}{lccrrr}
\toprule
Model
& GPUs
& TP
& Context length
& VideoQA cap
& Max. input tokens \\
\midrule
InternVL3.5-8B
& $1\times$ H200 & 1 & 40,960  & 128 & 33,442 \\

InternVL3.5-38B
& $2\times$ H200 & 2 & 40,960  & 128 & 33,442 \\

Qwen3-VL-2B
& $1\times$ H200 & 1 & 131,072 & 768 & 20,677 \\

Qwen3-VL-4B
& $1\times$ H200 & 1 & 131,072 & 768 & 16,164 \\

Qwen3-VL-8B
& $1\times$ H200 & 1 & 131,072 & 768 & 14,897 \\

Qwen3-VL-32B
& $1\times$ H200 & 1 & 131,072 & 768 & 20,011 \\

Qwen3.5-2B
& $1\times$ H200 & 1 & 131,072 & 768 & 14,908 \\

Qwen3.5-4B
& $1\times$ H200 & 1 & 131,072 & 768 & 14,908 \\

Qwen3.5-9B
& $1\times$ H200 & 1 & 131,072 & 768 & 16,783 \\

Qwen3.5-27B
& $1\times$ H200 & 1 & 131,072 & 768 & 18,642 \\
\bottomrule
\end{tabular}
\caption{Runtime configuration of the open-weight models. TP denotes
tensor parallelism; no data or pipeline parallelism is used. The last
column reports the maximum observed input length across the evaluated
requests. All observed inputs remain below the corresponding context limit;
therefore, no application-level input truncation is applied.}
\label{tab:open_runtime_configuration}
\end{table*}
\subsection{Fine-Grained Visual Detail Results}
\label{app:fine_grained_results}

\paragraph{\textrm{\textbf{CaptionQA shows model-dependent performance
across fine-grained visual tags.}}}
Figure~\ref{fig:accuracy_by_fine_grained_question_type} reports the
full model-wise accuracy across four fine-grained visual tags: object
identity, visual attributes, spatial relations, and actions. The
relative performance of CaptionQA and Direct VideoQA varies across
models and categories, indicating that caption-based representation
does not provide a uniform advantage for fine-grained visual
reasoning.

GPT-5.2 shows the clearest positive pattern: its 30-second CaptionQA
setting outperforms Direct VideoQA across all four tags. This result
indicates that GPT-5.2 can effectively use the fine-grained
information preserved in its generated captions. In contrast,
Direct VideoQA remains stronger than CaptionQA across the four tags
for Gemini 3 Flash, showing that strong native visual processing does
not necessarily benefit from conversion into textual memory.

The open-weight models exhibit similarly heterogeneous results.
Several Qwen models improve or remain competitive under CaptionQA for
particular object, attribute, spatial, or action subsets, but the
direction and magnitude of the differences vary by architecture and
tag. InternVL3.5 generally shows smaller or negative CaptionQA
differences. These variations are consistent with CaptionQA depending
jointly on what information is retained during caption generation and
how effectively the downstream model reasons over the resulting text.

Caption memory introduces a representational trade-off: text can
compactly support downstream QA but may omit small objects, subtle
attributes, spatial relations, or brief actions. Our analysis does not
show that textual retrieval is uniformly easier than visual reasoning.
This motivates retrieve-and-verify, which consults sparse visual
evidence to resolve details missing or ambiguous in captions.

\subsection{Full Evidence-Location Analysis}
\label{app:evidence_location}

\begin{figure*}[!t]
\centering
\includegraphics[
    page=1,
    width=0.95\textwidth,
    trim=0pt 30pt 0pt 50pt,
    clip
]{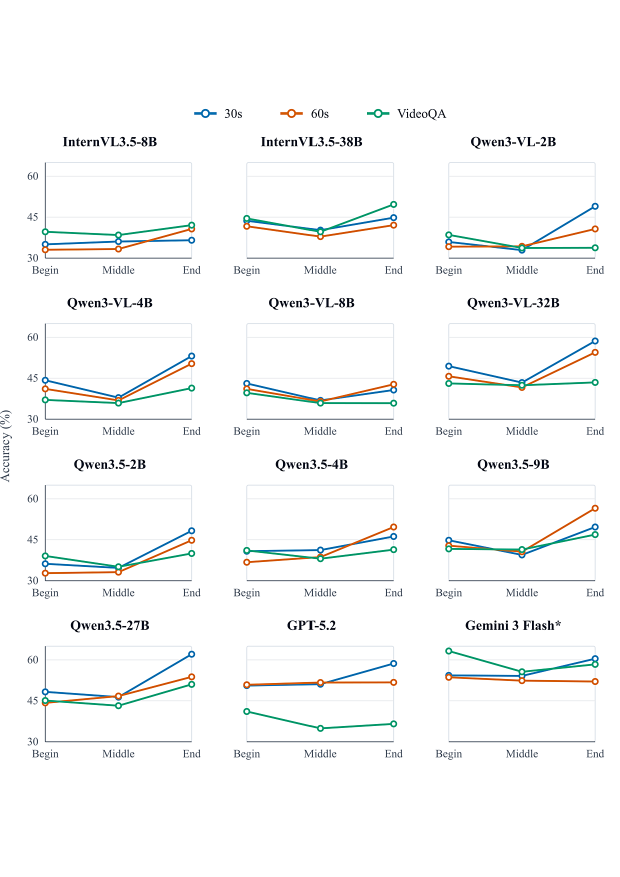}

\vspace{-5pt}
\caption{Full model-wise accuracy by evidence location. Accuracy is
grouped by whether the supporting evidence appears at the beginning
(0--20\%), middle (20--80\%), or end (80--100\%) of the video.
$^{\ast}$Gemini 3 Flash is computed on the 786-question VideoQA subset,
with CaptionQA evaluated on the same questions.}
\label{fig:accuracy_by_evidence_position1}
\end{figure*}

Figure~\ref{fig:accuracy_by_evidence_position1} provides the full
model-wise breakdown of QA accuracy by evidence location. Supporting
evidence is grouped into the beginning (0--20\%), middle
(20--80\%), and end (80--100\%) portions of each video. Several
models exhibit lower Direct VideoQA accuracy for middle-position
evidence, producing a U-shaped three-bin profile consistent with the
``lost-in-the-middle'' phenomenon. This pattern is visible for a
number of Qwen and InternVL variants, but is not shared uniformly
across all models.

CaptionQA produces a flatter evidence-location profile in several
cases. GPT-5.2 provides the clearest example: its Direct VideoQA
results contain a pronounced middle-position drop, whereas CaptionQA
keeps middle-evidence accuracy closer to its beginning- and
end-evidence performance. Qwen3.5-27B shows a related pattern under
the 30-second setting, which both improves overall accuracy and
reduces the observed middle-position drop. These comparisons show
that caption-based representation can be associated with less
middle-position degradation for particular models and settings.

A flatter profile does not necessarily imply higher absolute
accuracy. For some InternVL and smaller Qwen models, CaptionQA
reduces the variation across evidence-location bins while remaining
below Direct VideoQA overall. The two interfaces therefore exhibit
different error profiles in these cases. CaptionQA is additionally
limited by information omitted or imprecisely represented during
caption generation. The 30-second setting produces a flatter profile
than the 60-second setting for several models, which is consistent
with shorter caption windows retaining finer temporal detail, although
the pattern is not universal.

Overall, CaptionQA reduces middle-position degradation for some
models, but the pattern is not universal and does not establish a
causal retrieval mechanism. The differing profiles motivate the
retrieve-and-verify harness, in which captions provide candidate
temporal locations and sparse frames recover details that the
captions may omit.

\subsection{Duration Regimes and the VideoQA Frame Cap}
\label{app:duration_analysis}

At 1 FPS, the 768-frame VideoQA cap corresponds to approximately
12.8 minutes. We therefore divide the benchmark into
$\leq12$ minutes, $>12$--$\leq20$ minutes, and $>20$ minutes to test
whether CaptionQA's positive gap begins as soon as Direct VideoQA
reaches its frame limit. Table~\ref{tab:three_bin_duration} reports
question-weighted mean CaptionQA--Direct VideoQA gaps across the six
Qwen models used in the frame-aligned control.

If exceeding the frame cap alone caused the crossover, a positive
aligned gap should already appear in the intermediate
$>12$--$\leq20$-minute group. Instead, the aligned gap in this group
is zero at 30 seconds and negative at 60 seconds. Positive mean gaps
appear only in the strict $>20$-minute group, where CaptionQA exceeds
Direct VideoQA by 3.22 and 2.55 points at the two intervals.
Exceeding the 768-frame limit is therefore not sufficient to explain
the observed duration pattern.

One possible explanation is that longer videos contain more
temporally distributed or repeated evidence, which captions may
compress into shorter textual records. This interpretation remains
hypothetical: because we do not sweep multiple frame caps, the
$>20$-minute result should be treated as an empirical regime in
CapMem rather than a universal or cap-invariant threshold.

\begin{table}
\centering
\footnotesize
\setlength{\tabcolsep}{3.5pt}
\renewcommand{\arraystretch}{1.05}
\begin{tabular}{lrcccc}
\toprule
& & \multicolumn{2}{c}{Original}
& \multicolumn{2}{c}{Aligned} \\
\cmidrule(lr){3-4}\cmidrule(lr){5-6}
Duration & \#QA & 30s & 60s & 30s & 60s \\
\midrule
$\leq12$ min
& 148 & $-3.93$ & $-4.87$ & $-2.93$ & $-5.52$ \\
$>12$--$\leq20$ min
& 60 & $-2.52$ & $-4.72$ & $+0.00$ & $-4.72$ \\
$>20$ min
& 792 & $+4.62$ & $+2.22$ & $+3.22$ & $+2.55$ \\
\bottomrule
\end{tabular}
\caption{Mean CaptionQA--Direct VideoQA gaps (pp) across six Qwen
models by video duration.}
\vspace{-20pt}
\label{tab:three_bin_duration}
\end{table}

\subsection{Uncertainty Estimation}
\label{app:uncertainty_estimation}

Because frame alignment reduces CaptionQA's mean gains relative to
the full-coverage setting, we further assess whether the remaining
controlled effects are distinguishable from zero. We estimate
uncertainty over videos rather than treating the 1,000 questions as
independent observations, because questions associated with the same
video share visual content and may have correlated errors. The
analysis is based on the six Qwen models included in the frame-aligned
control.

Let $n_v$ denote the number of questions associated with video $v$.
For model $m$ and caption interval $s\in\{30,60\}$, we first compute
the paired video-level accuracy gap
\[
d_{v,m,s}
=
100\,
\frac{
C^{\mathrm{Cap}}_{v,m,s}
-
C^{\mathrm{Video}}_{v,m}
}{
n_v
},
\]
where $C^{\mathrm{Cap}}_{v,m,s}$ and
$C^{\mathrm{Video}}_{v,m}$ are the numbers of correct CaptionQA and
Direct VideoQA predictions on video $v$. We then average the gap
across the six models:
\[
\bar d_{v,s}
=
\frac{1}{6}\sum_{m=1}^{6}d_{v,m,s}.
\]

For a set of videos $\mathcal{V}$, the reported mean gap is weighted
by the number of questions associated with each video:
\[
\widehat{\Delta}_{s}
=
\frac{
\sum_{v\in\mathcal{V}} n_v\bar d_{v,s}
}{
\sum_{v\in\mathcal{V}} n_v
}.
\]
This weighting makes the point estimate equivalent to aggregating the
paired predictions over all questions and six models while retaining
the video as the unit of resampling.

We obtain 95\% confidence intervals using 10,000
duration-stratified video-cluster bootstrap samples. For the
full-benchmark analysis, we independently resample with replacement
the 28 videos in the $\leq20$-minute stratum and the 47 videos in the
$>20$-minute stratum, preserving the original number of videos in
each group. For duration-specific analyses, videos are resampled only
within the corresponding stratum. Whenever a video is sampled, all
of its questions and all six model results are retained as one
cluster. The question-weighted mean gap is recomputed for every
bootstrap sample, and the 2.5th and 97.5th percentiles of the 10,000
replicates form the reported 95\% confidence interval.

The resulting intervals are reported in
Table~\ref{tab:aligned_frame_uncertainty}. These intervals quantify
sampling uncertainty across the videos represented in CapMem; they
do not measure uncertainty over model families, since the set of six
Qwen models is held fixed.
\begin{table*}
\centering
\vspace{-60pt}
\scriptsize
\setlength{\tabcolsep}{4pt}
\renewcommand{\arraystretch}{1.05}
\begin{tabularx}{\textwidth}{
    >{\raggedright\arraybackslash}p{0.17\textwidth}
    >{\raggedright\arraybackslash}X
    >{\raggedright\arraybackslash}X
    >{\raggedright\arraybackslash}X
}
\toprule
Parameter
& Caption generation
& GPT-5.2 CaptionQA
& Gemini 3 Flash CaptionQA \\
\midrule
\multicolumn{4}{l}{\textit{Request configuration}} \\
\addlinespace[2pt]

Request unit
& One 30- or 60-second video window
& One question
& One question \\

Caption reuse
& Reused by all questions associated with the video
& --
& -- \\

Output limit
& 220 tokens
& 512 tokens$^\dagger$
& 64 tokens$^\dagger$ \\

Temperature
& 0
& 0
& 0 \\

Reasoning effort
& Minimal
& Minimal
& Minimal \\

Explicit thinking budget
& Not set
& Not set
& Not set \\

Concurrency
& 16
& 64
& 64 \\

Timeout
& 180 s
& 180 s
& 180 s \\

Candidate count
& Not explicitly set; only \texttt{choices[0]} is consumed
& Same
& Same \\

Seed
& Not set
& Not set
& Not set \\

Stop sequences
& Not set
& Not set
& Not set \\

Response format
& Plain text; JSON mode disabled
& Plain-text answer letter
& Plain-text answer letter \\

Safety settings
& Provider defaults
& Provider defaults
& Provider defaults \\

\midrule
\multicolumn{4}{l}{\textit{Retry and scoring policy}} \\
\addlinespace[2pt]

Maximum attempts
& \multicolumn{3}{p{0.74\textwidth}}{
One initial request plus up to eight retries (nine attempts in total).
} \\

Retried failures
& \multicolumn{3}{p{0.74\textwidth}}{
HTTP 408, 409, 429, 500, 502, 503, and 504 responses, together
with network exceptions.
} \\

Backoff
& \multicolumn{3}{p{0.74\textwidth}}{
Exponential backoff with random jitter; the base delay is capped
at 120 seconds.
} \\

Refusal or safety block
& \multicolumn{3}{p{0.74\textwidth}}{
Counted as incorrect when no answer in $\{A,B,C,D\}$ can be
extracted.
} \\

Empty or unparsable QA
& \multicolumn{3}{p{0.74\textwidth}}{
Counted as incorrect.
} \\

Semantic re-query
& \multicolumn{3}{p{0.74\textwidth}}{
No re-query is issued after a successful API response, even when
the response is empty, refused, or unparsable.
} \\

Empty caption
& Retained; downstream QA proceeds without a semantic retry
& --
& -- \\
\bottomrule
\end{tabularx}
\caption{Retained OpenRouter request and failure-handling configurations
for caption generation and CaptionQA. $^\dagger$ The two CaptionQA
output limits were recovered from truncation boundaries in the retained
results because the original complete launch commands were not
preserved.}
\vspace{-10pt}
\label{tab:closed_api_configuration}
\end{table*}

\begin{table*}

\centering
\small
\setlength{\tabcolsep}{7pt}
\renewcommand{\arraystretch}{1.05}
\begin{tabular}{llrr}
\toprule
Evaluation scope
& Task
& Selected frame identities
& Total frame inputs \\
\midrule
1,000 QA / 75 videos
& CaptionQA 30s       & 50,907 & 50,907 \\
& CaptionQA 60s       & 50,907 & 50,907 \\
& CaptionQA 30s + 60s & 50,907 & 101,814 \\
& Direct VideoQA      & 50,907 & 725,973 \\
\midrule
100 QA / 8 videos
& CaptionQA 30s       & 5,207 & 5,207 \\
& CaptionQA 60s       & 5,207 & 5,207 \\
& CaptionQA 30s + 60s & 5,207 & 10,414 \\
& Direct VideoQA      & 5,207 & 70,026 \\
\bottomrule
\end{tabular}
\caption{Frame accounting in the frame-aligned evaluations. Total
frame inputs count repeated use across caption intervals or
question-level Direct VideoQA evaluations.}
\label{tab:frame_accounting}
\vspace{-10pt}
\end{table*}
\subsection{Frame Accounting, Runtime, and Cost}
\label{app:compute_accounting}

\paragraph{Frame accounting.}
Let $T_v$ denote the number of 1-FPS frames in video $v$, $Q_v$ the
number of questions associated with that video, and $F_m$ the Direct
VideoQA frame cap of model $m$. In full-coverage CaptionQA, one
caption interval processes approximately $T_v$ frames once during
memory construction; evaluating both the 30- and 60-second settings
therefore processes approximately $2T_v$ frame inputs. Caption
generation is performed window by window and is not constrained by a
video-level frame cap.

Direct VideoQA uses at most $\min(T_v,F_m)$ selected frames for each
question. Its total number of question-level visual-frame inputs for
video $v$ is therefore
\[
N^{\mathrm{Video}}_{v,m}
=
Q_v\min(T_v,F_m).
\]
The cap $F_m$ is model-specific: it is 128 for InternVL3.5 and 768
for the Qwen models. In the frame-aligned Qwen experiment,
CaptionQA also uses $\min(T_v,768)$ frames, with exactly the same
frame identities as Direct VideoQA.

Table~\ref{tab:frame_accounting} reports aggregate frame counts in
the frame-aligned evaluations. Selected frame identities count unique
frames selected across the videos. Total frame inputs additionally
count repeated use across the two caption intervals or across
question-level Direct VideoQA evaluations.

CaptionQA processes each selected frame once when constructing the
memory for a given interval, after which downstream QA is text-only.
Direct VideoQA instead associates its selected visual frames with
each question-level evaluation. These counts describe the logical
visual inputs; implementation-level caching may reduce repeated
physical encoding.

\paragraph{Measured computation and cost.}
We measure Qwen3.5-2B and Qwen3.5-27B over all 75 videos and
1,000 questions on one NVIDIA H200. Gemini 3 Flash is measured in a
separate frame-aligned official-API timing run on the first
100 questions from eight videos. The CaptionQA measurements combine
the independent 30- and 60-second caption-generation and downstream
QA runs. Token counts include both memory construction and QA for
CaptionQA and all question-level requests for Direct VideoQA. Table~\ref{tab:measured_compute} summarizes the measured runtime,
H200-hours, API cost, and token usage. The reported cost analysis
covers the base CaptionQA pipeline and does not include the additional
retrieval and visual-verification calls used by the
retrieve-and-verify harness.

We report H200-hours rather than converting them into monetary cost
because the applicable local hourly rate was not preserved. The
Gemini timing experiment uses the official API and should not be
conflated with the 1,000-question OpenRouter main evaluation. Its
Direct VideoQA runtime includes rate-limit retries and failed
attempts, and the recorded API charge is therefore a lower bound.

Measured costs are model-dependent: both local Qwen models require
more time, GPU-hours, and input tokens for the two CaptionQA settings
than for one Direct VideoQA run, whereas Gemini requires fewer tokens,
less time, and lower API cost. The structural advantage is reuse:
caption memory is constructed once per video, while Direct VideoQA
reprocesses visual input for every question. Thus CaptionQA 
becomes increasingly amortized as more queries share the same memory.

\clearpage
\setcounter{dbltopnumber}{2}
\renewcommand{\dbltopfraction}{0.98}
\renewcommand{\textfraction}{0.02}
\setlength{\dbltextfloatsep}{6pt}

\begin{table*}[!t]
\centering
\scriptsize
\setlength{\tabcolsep}{3.5pt}
\renewcommand{\arraystretch}{1.05}
\begin{tabular}{lllrrrl}
\toprule
Model
& Scope
& Task
& Wall-clock
& H200-hours
& API cost
& Input / output tokens \\
\midrule

Qwen3.5-2B
& 1,000 QA
& CaptionQA 30s+60s
& 00:40:36.88
& 0.6769
& --
& 27,617,234 / 319,999 \\
Qwen3.5-2B
& 1,000 QA
& Direct VideoQA
& 00:15:03.74
& 0.2510
& --
& 13,525,676 / 3,009 \\
Qwen3.5-27B
& 1,000 QA
& CaptionQA 30s+60s
& 01:35:54
& 1.5983
& --
& 33,206,180 / 665,839 \\
Qwen3.5-27B
& 1,000 QA
& Direct VideoQA
& 00:33:01.63
& 0.5505
& --
& 13,525,676 / 2,001 \\
Gemini 3 Flash
& 100 QA
& CaptionQA 30s+60s
& 00:14:02.19
& --
& \$5.876472
& 11,815,823 / 29,344 \\
Gemini 3 Flash
& 100 QA
& Direct VideoQA
& $\approx$03:11:53
& --
& $\geq$\$7.101995
& 76,269,813 / 151 \\

\bottomrule
\end{tabular}

\caption{Measured computation, API cost, and token use. The Qwen
measurements cover the full benchmark on one H200, whereas the Gemini
measurements come from a separate official-API run on 100 questions.}
\label{tab:measured_compute}
\end{table*}

\clearpage
\twocolumn[{
\begin{minipage}{\textwidth}
\centering

\includegraphics[
    width=\textwidth
]{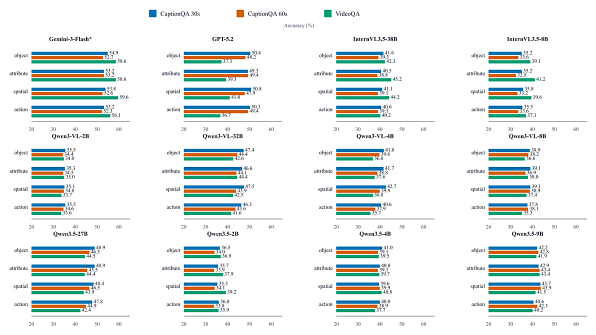}

\captionof{figure}{Full model-wise accuracy across fine-grained visual tags. Tags are defined from question-side fine-grained annotations and are not mutually exclusive. $^{*}$Gemini 3 Flash is computed on the 786-question VideoQA subset, with CaptionQA evaluated on the same questions.}
\label{fig:accuracy_by_fine_grained_question_type}

\vspace{0.5em}
\end{minipage}
}]

\section{Licensing and Release Details}
\label{app:licensing}

This appendix specifies the scope of the CapMem release and distinguishes CapMem-authored artifacts from third-party source media. We first describe the released artifacts and source-dataset terms, and then specify the licenses governing the CapMem-authored data and code.

\subsection{Release Composition and Source-Dataset Licensing}
\label{app:release_composition}

CapMem is constructed from 75 egocentric videos totaling 33.7 hours, selected from seven previously released datasets. The released benchmark consists only of independently authored CapMem annotations and associated metadata. It contains 1,000 multiple-choice QA records, each including a CapMem question identifier, source-dataset and source-video identifiers, the question and four answer options, the ground-truth answer, supporting evidence timestamps, domain and scenario labels, Ego/Exo and 4W1H categories, evidence-location labels, and multi-label fine-grained visual tags. We additionally release the official evaluation split, source-video manifest, evaluation prompts, evaluation, test scripts, and retrieve-and-verify scripts.

The release contains no source videos, frames, audio, or copied source-dataset annotations. Users must obtain the corresponding media from the original providers and comply with their respective licenses, access conditions, attribution requirements, and privacy restrictions. Table~\ref{tab:source_video_composition} lists the source-dataset composition and video identifiers, while Table~\ref{tab:source_licenses} summarizes the governing source-media terms and their treatment in the CapMem release.
\begin{table*}[p]
\centering
\scriptsize
\setlength{\tabcolsep}{4pt}
\renewcommand{\arraystretch}{1.08}
\begin{tabularx}{\textwidth}{
    l
    r
    r
    r
    >{\raggedright\arraybackslash}X
}
\toprule
Source dataset & \#Videos & \#QA & Hours & Source video identifiers  \\
\midrule

Ego4D
& 56 & 742 & 22.489 &
\path{a887acbe-efd5-48e2-ad8b-8e22a3c654d2.mp4},
\path{e303a071-7d0b-40e7-b8df-dea6e79ad7e8.mp4},
\path{23164be8-8b5e-492f-8417-a56f4f44d9db.mp4},
\path{01111831-9107-43c4-bf0e-6b26e9b32a2b.mp4},
\path{f71187bc-cead-407d-ad14-62812993b6a3.mp4},
\path{3906d25a-d0a4-4a7c-8906-a87cea106c66.mp4},
\path{80c74a5b-81e0-4bc1-bc09-2f146351fe70.mp4},
\path{962fc84c-470e-430b-beef-ec892f936f5c.mp4},
\path{f3b1d2c0-728b-40c4-bc8f-236b7a558dcc.mp4},
\path{406c7c13-4390-4c19-99c6-68861245da0d.mp4},
\path{6f3107a6-bd6a-4761-a44a-f53c40b47f7d.mp4},
\path{b0a30641-ad93-4d6a-92a9-f277c78f1aa5.mp4},
\path{f60c7255-97fb-4f8b-a12d-f8f805423bfe.mp4},
\path{267aa640-f3d0-4288-96b2-6ecf32b23198.mp4},
\path{b79c342b-db1e-407e-958a-e1585c6c3ddd.mp4},
\path{0075dfd4-03b2-48c3-bde1-d7851bc8e367.mp4},
\path{3b3ecf0c-ee8d-4518-bf1c-3301ba44c743.mp4},
\path{9155f5b9-6235-484f-990d-fbd3010a791e.mp4},
\path{1680e3e9-d3fc-4903-94d8-aa8e74016d74.mp4},
\path{320f0ec0-0e73-43aa-8375-1a030793eb1e.mp4},
\path{f41432c6-1ef3-48f6-8646-887ab70cb031.mp4},
\path{07536df6-9813-42d9-93df-35835407d8ee.mp4},
\path{114c1877-5b5b-4ca3-9399-bd535d55669a.mp4},
\path{9bbaa1a1-2d12-4e85-b9ce-fc5a738bf38a.mp4},
\path{f8c36c51-1cca-4ed3-81db-c190b72d5c49.mp4},
\path{6099730e-7a7c-41ce-8266-83a9f00cacec.mp4},
\path{ce19b572-7341-426a-8c2f-fe067344c98d.mp4},
\path{7a07a236-aaba-48ee-90a8-598fb6626cc2.mp4},
\path{820d4b9e-6fba-472f-aa45-25b27241d4af.mp4},
\path{e99a32fa-19c9-41e6-be66-37869bd26583.mp4},
\path{6b6d73b9-e92c-4698-86df-bbe687a9e95e.mp4},
\path{11586867-f4d9-4341-9532-4de21a47585d.mp4},
\path{a6f99fa0-6ad4-4b1c-b7bd-68d69e708736.mp4},
\path{a8197601-7474-4cb0-8d08-e8a8febcecfc.mp4},
\path{02ca6897-bf08-4cfb-bbc7-7810dbf5cca2.mp4},
\path{1ebbf490-8a85-46b9-99c9-cd0f7e3b634b.mp4},
\path{60bfacbb-063f-4e3f-bc74-fe993c7a521a.mp4},
\path{8e84c762-b0f9-4a7d-982e-010f27999435.mp4},
\path{bc16dcef-d666-42c8-8a83-6babdf0e045f.mp4},
\path{a01446fe-4495-4424-b729-b3f5661140e5.mp4},
\path{cb87d2ca-dc4f-4e1c-a13f-a8fe7dc952e9.mp4},
\path{e750d46b-2c3a-468e-88db-4169a9d65a9a.mp4},
\path{b7420366-105d-4583-b1d2-ecfbf2166733.mp4},
\path{def26f88-2696-4b70-a545-33ad481ca696.mp4},
\path{fdf36515-a08a-4575-9a79-65d0ec260cc7.mp4},
\path{7c925095-c62a-4a7a-a2ca-118ca3f0176a.mp4},
\path{dced4894-d13a-4222-941b-5e4ecf1e3299.mp4},
\path{f22be6c4-8951-47f3-a8f4-d1a61912b4ff.mp4},
\path{af215e0f-56c7-4356-9aad-a3a5fb7bbc4f.mp4},
\path{033520cd-e7e4-4649-b551-83bb451a4802.mp4},
\path{41309ada-0a81-47b2-bdd1-453a178db18a.mp4},
\path{3f8dbb4f-ad5e-4d54-bf95-3ddbe968a85d.mp4},
\path{d51cd3c2-dc8a-42ad-90c7-a130f6481b7d.mp4},
\path{51ab3f5f-f17f-4726-8f2c-b8e504e47377.mp4},
\path{542bd8e2-37a1-49be-8f4c-0fedeb1cc9f1.mp4},
\path{5c520abb-dee8-4b65-9a2a-3112a42e79a9.mp4}
\\

\addlinespace
EPIC-KITCHENS
& 4 & 61 & 1.967 &
\path{EPIC-KITCHENS_P01_03.MP4},
\path{EPIC-KITCHENS_P01_05.MP4},
\path{EPIC-KITCHENS_P01_105.MP4},
\path{EPIC-KITCHENS_P01_09.MP4}
\\

\addlinespace
EgoLife
& 5 & 89 & 3.328 &
\nolinkurl{DAY1_A1_JAKE_11493000--DAY1_A1_JAKE_12290000},
  \nolinkurl{DAY2_A3_TASHA_11200000},
  \nolinkurl{DAY5_A3_TASHA_15193000--DAY5_A3_TASHA_16160000},
  \nolinkurl{DAY6_A1_JAKE_12420000},
  \nolinkurl{DAY7_A4_LUCIA_14243000}
\\

\addlinespace
CASTLE2024
& 3 & 51 & 3.000 &
\path{CASTLE2024_day1_Allie_12.mp4},
\path{CASTLE2024_day1_Bjorn_09.mp4},
\path{CASTLE2024_day2_Allie_15.mp4}
\\

\addlinespace
EgoPet
& 5 & 39 & 2.284 &
\path{EgoPet_edited_3a628da0633ba2685a871f84e840129e7885d0571e00377024f25255d5638ef6_segment_25.mp4},
\path{EgoPet_edited_f0596f2b564265bb5f34dc2042148a61fbc62e3bba7089c79f4c40ccb12e40d3_segment_011.mp4},
\path{EgoPet_edited_2601ab6126bb4fdc6aa41e307134b48a14b317228b4bb1e566adb7886e762899_segment_002.mp4},
\path{EgoPet_edited_ca553b127e28b214fc5130f384ca5341912318ef4e23d6e4935388451bdf819a_segment_003.mp4},
\path{EgoPet_edited_e639364fb4e2e34c2c4dab43cba2c1809cf8f0de12dad84b80cc849b6dc34390_segment_1.mp4}
\\

\addlinespace
ENIGMA-51
& 1 & 4 & 0.278 &
\path{ENIGMA51_141.mp4}
\\

\addlinespace

HoloAssist
& 1
& 14
& 0.371
& \nolinkurl{z015-june-17-22-rashult_assemble}
\\

\midrule
Total & 75 & 1,000 & 33.717 & --- \\
\bottomrule
\end{tabularx}

\vspace{2pt}
\begin{minipage}{\textwidth}
\scriptsize
\end{minipage}

\caption{Source-dataset composition and source-video identifiers used by CapMem.}
\vspace{-10pt}
\label{tab:source_video_composition}
\end{table*}

\begin{table*}[t]
\vspace{-200pt}
\centering
\small
\setlength{\tabcolsep}{6pt}
\renewcommand{\arraystretch}{1.1}
\begin{tabularx}{\textwidth}{
    l
    >{\raggedright\arraybackslash}X
    >{\raggedright\arraybackslash}X
}
\toprule
Source dataset & Source-media license or access terms
& CapMem release treatment \\
\midrule

Ego4D
& Ego4D Dataset License Agreement; access requires acceptance of
the official agreement.
& CapMem releases source identifiers and independently authored
annotations only; no source media or copied source annotations. \\

EPIC-KITCHENS
& Creative Commons Attribution-NonCommercial 4.0 International
(CC BY-NC 4.0).
& Source media are excluded; users obtain them from the original
provider and comply with attribution and non-commercial-use terms. \\

EgoLife
& S-Lab License 1.0; non-commercial use subject to its attribution
and redistribution conditions.
& Source media are excluded. \\

CASTLE2024
& Creative Commons Attribution-NonCommercial-ShareAlike 4.0
International (CC BY-NC-SA 4.0), together with the project Terms
of Use.
& CapMem releases no source media or copied annotations; users remain
subject to the original access and non-commercial-use terms. \\

EgoPet
& Creative Commons Attribution-NonCommercial 4.0 International
(CC BY-NC 4.0).
& Source media are excluded; CapMem releases only source identifiers
and independently authored annotations. \\

ENIGMA-51
& No explicit public dataset license is stated on the official
project page.
& Source media are excluded.  \\

HoloAssist
& Community Data License Agreement--Permissive 2.0
(CDLA-Permissive-2.0).
& Source media are excluded; CapMem releases only source identifiers
and independently authored annotations. \\

\bottomrule
\end{tabularx}
\caption{Licensing and release treatment of the source media represented
in CapMem.}
\label{tab:source_licenses}
\end{table*}
\subsection{CapMem License}\label{app:capmem_license}

The independently authored CapMem QA annotations and associated metadata are released under the Creative Commons Attribution-NonCommercial 4.0 International license (CC BY-NC 4.0). This release includes the questions and answer options, ground-truth answers, evidence timestamps, taxonomy and visual-tag labels, the official evaluation split, source-video manifest, and evaluation prompts. These materials may be shared and adapted for non-commercial purposes with appropriate attribution.

The evaluation, test scripts, and retrieve-and-verify scripts are released separately under the Apache License 2.0. The corresponding license texts and attribution instructions are included in the release as \texttt{LICENSE-DATA} and \texttt{LICENSE-CODE}.

These licenses cover only CapMem-authored artifacts. Source media
and source-dataset annotations are not redistributed and remain
governed by their original terms; users must obtain the media from
the original providers.





\end{document}